\pdfoutput=1

\documentclass[11pt]{article}

\usepackage[table,xcdraw]{xcolor}

\usepackage{acl}

\usepackage{times}
\usepackage{latexsym}

\usepackage[T1]{fontenc}

\usepackage[utf8]{inputenc}

\usepackage{microtype}

\usepackage{color}
\usepackage{soul}
\usepackage{amsmath}
\usepackage{amsfonts}
\usepackage{amssymb}
\usepackage{graphicx}
\usepackage{enumitem}
\usepackage{booktabs}
\usepackage{multirow}
\usepackage{array}
\usepackage{float}
\usepackage{arydshln}
\usepackage{adjustbox}
\usepackage{multicol}
\usepackage{mathabx}
\usepackage{soul}
\usepackage{inconsolata}
\usepackage{xspace}
\usepackage{tcolorbox}
\usepackage[multiple]{footmisc}
\usepackage[section]{placeins}
\usepackage[normalem]{ulem}
\usepackage[framemethod=TikZ]{mdframed}

\newmdenv[
  linecolor=white!50!black!50,       
  linewidth=0.5pt,         
  roundcorner=5pt,      
  innertopmargin=10pt,   
  innerbottommargin=10pt,
  innerleftmargin=10pt,  
  innerrightmargin=10pt, 
  backgroundcolor=white, 
  font=\footnotesize            
]{externaldoc}

\newcommand*{\datatool}{\textsc{Quintd}\xspace}
\newcommand*{\benchmark}{\textsc{Quintd}-1\xspace}
\newcommand*{\gptmetric}{$\mathcal{E}_{\text{gpt}}$\xspace}
\newcommand*{\humanmetric}{$\mathcal{E}_{\text{hum}}$\xspace}

\setlist[itemize]{itemsep=0.02cm,topsep=0.2cm}
\setlist[enumerate]{itemsep=0.02cm,topsep=0.2cm}

\definecolor{errred}{HTML}{C9404F}
\definecolor{erryellow}{HTML}{C9AB40}
\definecolor{errpurple}{HTML}{954CB0}
\definecolor{errgrey}{HTML}{A0A0A0}

\newcommand{\errinc}[1]{{\leavevmode\color{errred} \ul{\textbf{#1\textsuperscript{I}}}}}
\newcommand{\errmis}[1]{{\leavevmode\color{erryellow} \ul{\textbf{#1\textsuperscript{M}}}}}
\newcommand{\errnc}[1]{{\leavevmode\color{errpurple} \ul{\textbf{#1\textsuperscript{NC}}}}}
\newcommand{\errother}[1]{{\leavevmode\color{errgrey} \ul{\textbf{#1\textsuperscript{O}}}}}

\newtcolorbox{verbatimbox}{
  colback=black!4, 
  colframe=black!4, 
  arc=1pt, 
  boxrule=0.5pt, 
  fontupper=\ttfamily 
}

\makeatletter
\def\adl@drawiv#1#2#3{%
        \hskip.5\tabcolsep
        \xleaders#3{#2.5\@tempdimb #1{1}#2.5\@tempdimb}%
                #2\z@ plus1fil minus1fil\relax
        \hskip.5\tabcolsep}
\newcommand{\cdashlinelr}[1]{%
  \noalign{\vskip\aboverulesep
           \global\let\@dashdrawstore\adl@draw
           \global\let\adl@draw\adl@drawiv}
  \cdashline{#1}
  \noalign{\global\let\adl@draw\@dashdrawstore
           \vskip\belowrulesep}}
\makeatother

\usepackage[normalem]{ulem}
\def\ODdel#1{\bgroup\markoverwith{\textcolor{purple!60}{\rule[0.4ex]{2pt}{3pt}}}\ULon{#1}}

\title{Beyond Traditional Benchmarks:\\ Analyzing Behaviors of Open LLMs on Data-to-Text Generation}

\author{Zdeněk Kasner \and Ondřej Dušek \\
  Charles University, Faculty of Mathematics and Physics\\
  Institute of Formal and Applied Linguistics \\
  Prague, Czech Republic \\
  \texttt{\{kasner,odusek\}@ufal.mff.cuni.cz}
}

\begin{document}
\maketitle

\begin{abstract}
  We analyze the behaviors of open large language models (LLMs) on the task of data-to-text (D2T) generation, i.e., generating coherent and relevant text from structured data. To avoid the issue of LLM training data contamination with standard benchmarks, we design \datatool{} -- a tool for collecting novel structured data records from public APIs. We find that open LLMs (Llama 2, Mistral, and Zephyr) can generate fluent and coherent texts in zero-shot settings from data in common formats collected with \datatool. However, we show that the semantic accuracy of the outputs is a major issue: both according to human annotators and our reference-free metric based on GPT-4, more than 80\% of the outputs of open LLMs contain at least one semantic error. We publicly release the code, data, and model outputs.\footnote{\url{https://d2t-llm.github.io/}}
\end{abstract}

\section{Introduction}

\begin{figure}[t]
  \centering
  \includegraphics[width=\columnwidth]{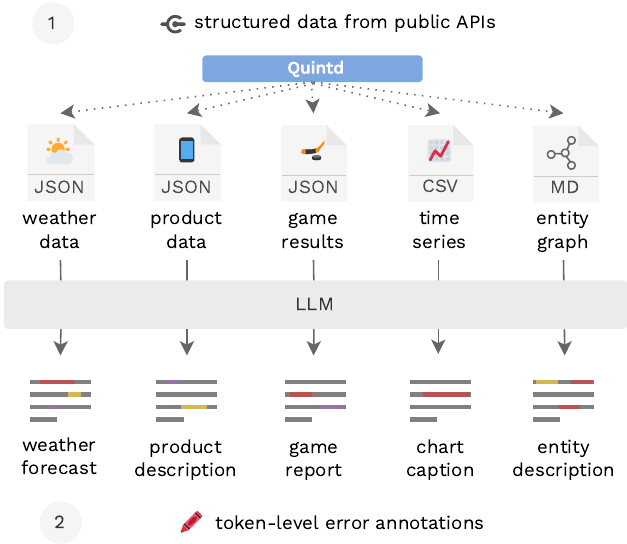}
  \caption{To benchmark LLMs, we download unlabeled structured data from public APIs and prompt LLMs to generate texts based on the data. We annotate semantic errors in the outputs using reference-free metrics.}\label{fig:teaser}
\end{figure}

\begin{table*}[ht]
  \small
  \centering
  \begin{tabular}{llllll} \toprule                                                                                                                                                      \\
    \textbf{Task Id}     & \textbf{Domain} & \textbf{Task Description}                               & \textbf{Source}                                   & \textbf{Format} \\ \midrule
    \texttt{openweather} & Weather         & Generating a weather forecast from weather data.        & \href{https://openweathermap.org}{OpenWeather}    & JSON            \\
    \texttt{gsmarena}    & Technology      & Describing a product based on its attributes.           & \href{https://www.gsmarena.com}{GSMArena}         & JSON            \\
    \texttt{ice\_hockey} & Sport           & Describing an outcome of an ice-hockey game.            & \href{https://rapidapi.com}{RapidAPI}             & JSON            \\
    \texttt{owid}        & Health          & Generating a caption for a time series.                 & \href{https://ourworldindata.org}{OurWorldInData} & CSV             \\
    \texttt{wikidata}    & World facts     & Describing entities and relations in a knowledge graph. & \href{https://wikidata.org}{Wikidata}             & Markdown        \\\bottomrule
  \end{tabular}
  \caption{The domains and tasks included in the \datatool data collection tool we use for testing D2T generation with LLMs. In our experiments, we download 100 development and 100 test examples of input data for each domain.}
  \label{tab:data}
\end{table*}

Large language models (LLMs; \citealp{Ouyang2022TrainingLM,touvron2023llama,touvronLlamaOpenFoundation2023,jiangMistral7B2023,tunstallZephyrDirectDistillation2023}) have already left a mark in many areas of natural language processing (NLP). Surprisingly, their applicability to the task of data-to-text (D2T) generation \cite{reiter1997building,gatt2018survey} remains underexplored, with limited evaluation on a handful of well-established benchmarks only \cite{axelssonUsingLargeLanguage2023,yuanEvaluatingGenerativeModels2023}. Generating text from structured data is arguably challenging for LLMs, given the specifics of D2T generation, such as long inputs, complex non-linear structure, and strict requirements on semantic accuracy. However, a more significant issue is the lack of testing grounds. The current D2T generation benchmarks are not only getting saturated \cite{vanmiltenburgBarriersEnablingFactors2023}, but also promote optimization towards traditional reference-based evaluation metrics, which were shown to correlate poorly with human judgment \cite{gehrmann2022repairing,vanderleeHumanEvaluationAutomatically2021,novikovaWhyWeNeed2017}. When it comes to the models, using closed LLMs \cite{openai2023gpt4,chatgpt} is increasingly considered a bad research practice due to its non-reproducibility \cite{rogers2023closed,chen2023chatgpt}. On top of that, contamination of LLM training data with standard benchmarks further restricts the space for experiments \cite{golchin2023time,aiyappa-etal-2023-trust,balloccu2024leak}.

In this paper, we propose an approach that allows us to analyze model behavior in D2T generation on novel, real-world structured data records with reference-free evaluation metrics. We begin by realizing that \textit{unlabeled data are plentiful}. To leverage the data for our experiments, we introduce \datatool\footnote{\underline{Q}uintet of \underline{U}nlabeled \underline{I}nputs for \underline{N}atural \underline{T}asks in \underline{D}ata-to-text, pronounced as ``quintet''} -- a tool for collecting structured data from five domains in standard formats: JSON,
CSV,
and Markdown.
We choose the domains so that the data can be directly used as input for five distinct D2T generation tasks. Our tasks include generating weather forecasts, sports reports, product descriptions, chart captions, and entity descriptions (see \autoref{tab:data}).
Next, we collect a set of 1,000 inputs with \datatool and use the inputs as an ad-hoc benchmark (called \benchmark) for testing the abilities of LLMs for D2T generation. We assume that the data formats in \benchmark are common in the LLMs' pretraining corpora, so we specify the task using instructions instead of standard finetuning with human-written outputs, capitalizing on the zero-shot abilities of instruction-tuned LLMs (§\ref{sec:data}).

We push towards better reproducibility by \textit{focusing on open LLMs}, which -- apart from being more accessible -- also achieve increasingly better results across tasks \cite{zheng2023judging,open-llm-leaderboard}. For our experiments, we use three open LLMs with 7B parameters: Llama 2 \cite{touvronLlamaOpenFoundation2023,llama-2-7b-32k}, Mistral \cite{jiangMistral7B2023}, and Zephyr \cite{tunstallZephyrDirectDistillation2023}. We also use GPT-3.5 \cite{chatgpt} as a closed model baseline for the final experiments. Given the behavioral nature of the experiments with LLMs \cite{holtzmanGenerativeModelsComplex2023}, we put emphasis on reporting model behavior throughout the process  (§\ref{sec:experiments}).

Another piece of the puzzle is \textit{reference-free evaluation}: using the input data as a ground for comparison instead of reference outputs (§\ref{sec:eval}). We focus on identifying semantic errors in the model outputs, i.e., the information that is not supported by the input data. We use two separate evaluation methods: manual annotations from human crowdworkers \cite{vanderleeHumanEvaluationAutomatically2021} and a custom automatic metric based on GPT-4 \cite{liuGEvalNLGEvaluation2023,chiang-lee-2023-large,kocmiGEMBAMQMDetectingTranslation2023}. We annotate the errors on the level of individual words, getting fine-grained annotations of error spans in several categories \cite{thomsonGoldStandardMethodology2020,thomson2023evaluating}.

Based on our results, we provide general recommendations for D2T generation with open LLMs across tasks and formats (§\ref{sec:discussion}). Our main findings are as follows:
\begin{itemize}
  \item \textbf{Open LLMs can generate fluent outputs from structured data} in common formats under zero-shot settings.
  \item \textbf{Semantic accuracy is a major obstacle}: both human annotators and GPT-4-based metric report that over 80\% of outputs of open LLMs on our data contain a semantic error.
  \item \textbf{Long data inputs cause practical issues}, including the need for long-context models, increased GPU memory requirements, and unavailability of few-shot approaches.
  \item \textbf{Outputs can be empirically improved by following several rules-of-thumb} for preprocessing the model input, such as including units, removing unnecessary fields, or prefixing the model answer.
\end{itemize}

\section{Reference-Free D2T Generation}
\label{sec:data}

\subsection{Data Collection Tool}
\label{sec:data_collection}
We introduce \datatool{},\footnote{\url{https://github.com/kasnerz/quintd}} a tool for collecting ad-hoc test sets using public APIs in five different domains.
Our main reasons for departing from the traditional scheme of benchmarking on well-established datasets are:
\begin{enumerate}
  \item Any published test sets may be potentially included in the training data of LLMs.
  \item Public sources of structured data offer enough resources for creating ad-hoc test sets.
  \item Without human references, our data collection scheme is lightweight and replicable.
\end{enumerate}

Given the available public sources of data, we settled on the five tasks which are described in \autoref{tab:data} (see Appendix \ref{app:data} for more details). The tasks are based on structured data in common formats: JSON, CSV, and Markdown.

\begin{figure}[t]
  \centering
  \small
  \textbf{Prompt}
  \begin{verbatimbox}
    Based on the given data:

    \`{}\`{}\`{}

    \{data\}

    \`{}\`{}\`{}

    Your task is to write a brief, fluent, and
    coherent single-paragraph \{output\_type\} in natural
    language. The text should be balanced and neutral.
    Make sure that all the facts mentioned in the text
    can be derived from the input data, do *not* add
    any extra information.
  \end{verbatimbox}
  \textbf{Output prefix}
  \begin{verbatimbox}
    Sure! Here is the \{output\_type\}:

    "
  \end{verbatimbox}
  \caption{The prompt $\mathcal{P}$ and the model output prefix we used for the experiments in this paper. \texttt{data} is filled with the data record  $x$ and \texttt{output\_type} is filled accordingly for each domain $\mathcal{D}$ (see Table~\ref{tab:data} and Table~\ref{tab:types} in the Appendix).}
  \label{fig:prompt}
\end{figure}

\subsection{\benchmark Dataset}
\label{sec:dataset}
Using \datatool{}, we collected the dataset for our experiments in this paper (\datatool{}-1). The dataset contains 500 examples in the development set and 500 examples in the test set (100 examples per domain for each split).
We downloaded the data between November 2023 and January 2024. Note that the dataset contains only \textbf{unlabeled} data without any reference outputs (e.g., weather data, but not a textual weather forecast), so the outputs need to be evaluated using reference-free metrics. New versions of the benchmark can be easily generated with our \datatool{} tool.

\subsection{Task Definition}
Each example in \benchmark consists of a structured data record $x$ from a domain $\mathcal{D} \in $ \{\texttt{openweather}, \texttt{gsmarena}, \texttt{ice\_hockey}, \texttt{owid}, \texttt{wikidata}\}. Given $x$ and a prompt $\mathcal{P}$, the goal is to generate natural language output $y$ faithful to the data $x$, according to the instructions in the prompt $\mathcal{P}$ (see \autoref{fig:prompt}).



\section{Experiments}
\label{sec:experiments}
\subsection{Experimental Process}
\label{sec:process}
Our goal is to avoid extensive data preprocessing and prompt engineering since these steps could harm the reproducibility and generalizability of our experiments. With this goal in mind, we decided to use the same prompt template $\mathcal{P}$ for all the domains and models.

For a set of preliminary experiments, we first wrote down the initial version of the prompt and used the data without further preprocessing.
We then iteratively improved our experimental setup by observing outputs on the development set.
In §\ref{sec:observations}, we describe all the observations and modifications we made before generating the final outputs on the test set.

\begin{figure*}[ht]
  \centering
  \includegraphics[width=\textwidth]{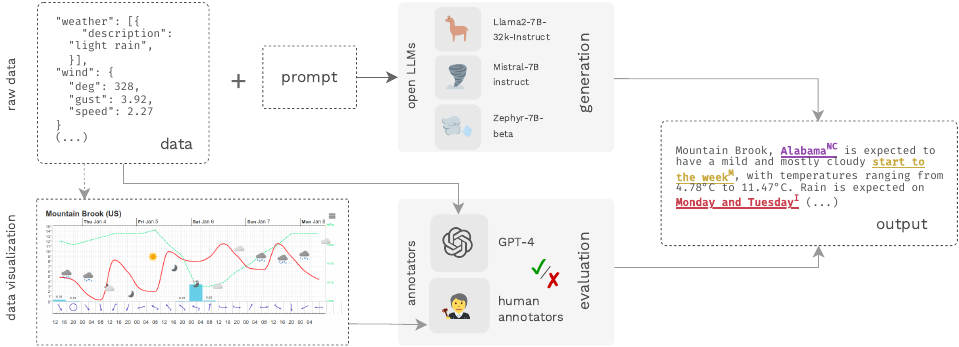}
  \caption{Our experimental setup. We first generate the outputs using LLMs that are given raw data and a task-specific prompt. We annotate the word-level semantic errors in the LLM outputs with (a) an automatic metric based on GPT-4 that matches the output to the raw data, and (b) human annotators, who annotate the errors in the output given the data visualization.}\label{fig:process}
\end{figure*}

\subsection{Models}
\label{sec:models}
For our experiments, we selected the following LLMs available under an open license:

\begin{itemize}
  \item \textbf{Llama 2} \cite{touvron2023llama,llama-2-7b-32k},\\ \texttt{\small together\-computer\-/Llama-2-7B\--32K-Instruct}
  \item \textbf{Mistral} \cite{jiangMistral7B2023},\\ \texttt{\small mistralai/Mistral-7B-Instruct-v0.1}
  \item \textbf{Zephyr}  \cite{tunstallZephyrDirectDistillation2023}. \\ \texttt{\small HuggingFaceH4/zephyr-7b-beta}
\end{itemize}

The models are instruction-tuned, operate with 32k context, and perform well on recent benchmarks \cite{open-llm-leaderboard}. All the models have 7B parameters and thus fit on a single NVIDIA A40 (48G VRAM) in 16-bit precision.  The models are available through HuggingFace \cite{wolf2019huggingface}.

We accessed the models via an API provided by the \texttt{text-generation-webui} framework\footnote{\url{https://github.com/oobabooga/text-generation-webui}} running locally.
For the final experiments, we also included GPT-3.5 (\texttt{gpt-3.5-turbo-1106}) accessed through the OpenAI API \cite{chatgpt}.\footnote{We only included GPT-3.5 in our final experiments as proprietary models were not our focus. We did not include GPT-4 since we use the same model for evaluation (see §\ref{sec:gpt4eval}) and LLMs tend to be biased towards their own outputs \cite{kooBenchmarkingCognitiveBiases2023,stureborgLargeLanguageModels2024}.}

\subsection{Observations from Preliminary Experiments}
\label{sec:observations}
During development, we made several observations which we took into account for our final experimental setup:

\paragraph{Any input field may appear in the output.} The models do not always select the most relevant fields for the given output. For example, we observed that the models commonly mention identifiers, timestamps, files, and other metadata, leading to unnatural outputs. To mitigate these issues, we manually picked irrelevant fields and filtered them out from the input.

\paragraph{Units need to be specified explicitly.} If the units are not specified in the data record, the models tend to resort to their best guess. This may go unnoticed if the unit is evident from the context (e.g., the model will usually not report the temperature in Fahrenheit instead of Celsius), but it may get problematic if the value is ambiguous (e.g., wind speed in km/h versus m/s). Therefore, we explicitly add units to all data records where appropriate.

\paragraph{Understandable field names are enough.} On the flip side, we decided not to add extra descriptions to field names in the data if the field was understandable from its name (e.g., \texttt{homeTeam} or \texttt{dimensions}). As discussed by \citet{kasner2023mind}, pretrained models interpret field names correctly as long as they are human-readable. We only include chart metadata for the CSV files in the \texttt{owid} domain.

\paragraph{Long inputs can be troublesome.} The inputs in some domains can easily get longer than 10-20k tokens. This issue is amplified by the fact that the evaluated LLMs tokenize numbers into individual digits. To accommodate for the long inputs, we picked models that accept up to 32k tokens.\footnote{For this reason, we use Llama-2-7B-32k with 32k token context \cite{llama-2-7b-32k} instead of the official Llama-2-7B-Instruct, which only supports 4k context \cite{touvronLlamaOpenFoundation2023}.} However, with long inputs, the GPU memory consumption also gets considerably higher, so we needed to downsample the data in \texttt{owid} and \texttt{openweather} to keep their length under \textasciitilde 8k tokens.

\paragraph{Few-shot experiments are infeasible.} Due to the above context-length limitations, we were not able to run few-shot experiments since we could not robustly fit an additional ($x_\text{example}$, $y_\text{example}$) pair in the prompt. We attempted to include only $y_\text{example}$ (making the setup ``half-shot''), but we observed that the models then used entities from the example (unrelated to the actual input) in their outputs. Therefore, we decided to leave this line of experiments for future work (see §\ref{sec:future} for discussion).

\paragraph{Deterministic decoding and sampling are on par.} In our preliminary experiments, we observed a roughly similar output quality for both deterministic decoding and sampling.\footnote{We used the \texttt{text-generation-webui} default decoding parameters: \texttt{temperature}=0.7, \texttt{top\_p}=0.9, and \texttt{top\_k}=20.} For the final experiments, we decided to use deterministic decoding, which is non-parametric and conceptually more suitable for D2T generation.

\paragraph{Prefixing the output makes parsing easier.} Even with variations of a \textit{``generate only the output''} instruction appended to the prompt, the models (especially Llama 2) tended to first confirm the request. For that reason, we decided to prefix the input for all the models with \textit{``Sure! Here is the \{output\_type\}: "''}. The opening quote at the end of the prefix allowed us to robustly parse the text simply by stripping the closing quote from the model output.

\paragraph{The outputs are fluent but inaccurate.} We observed that the vast majority of model outputs were grammatically and stylistically correct, capturing the output type specified in the prompt. However, we also noticed that the outputs contained many semantic errors (even after emphasizing the focus on semantic accuracy in the prompt, see \autoref{fig:prompt}). This observation led us to evaluate the model outputs using word-level annotations focused on semantic accuracy errors (see §\ref{sec:eval}).

\subsection{Final Experiments}
\label{sec:basic}

Taking the observations in §\ref{sec:observations} into account, we proceeded to generate the outputs on the test set of \benchmark for word-level error analysis. We first preprocessed the data as mentioned: we stripped out unnecessary fields, added units, and downsampled the data to fit the context. For all the models mentioned in §\ref{sec:models}, we used the prompt in \autoref{fig:prompt} and deterministic decoding with a maximum length of 512 tokens.

For comparison, we also generated outputs for the same inputs and identical prompts with GPT-3.5. Note that even though we fixed the temperature and seed to $0$, the rest of the decoding parameters are inaccessible to us and may differ from the parameters we used for the open models.

\section{Evaluation}
\label{sec:eval}
For evaluation, we focus on identifying \emph{semantic errors} in model outputs. We compare the generated texts to the input data, looking for parts of texts that are not faithful to the input data. We annotate the errors on the word level, considering all the words in the output text as potential sources of errors.

We use two complementary referenceless evaluation methods:
\begin{itemize}
  \item \humanmetric{}: \textbf{human evaluation} based on crowdsourcing (§\ref{sec:humaneval}),
  \item \gptmetric{}: \textbf{an automatic metric} based on GPT-4 (§\ref{sec:gpt4eval}).
\end{itemize}

The methods use similar instructions and produce outputs with equivalent format. The main idea is to compensate for the shortcomings of each approach: while human evaluation is costly (about ten times more expensive than automatic evaluation), using only an automatic metric based on a closed LLM would make the evaluation potentially non-reproducible and biased \cite{kocmiGEMBAMQMDetectingTranslation2023,wangLargeLanguageModels2023}. 
Reporting the results of both methods should hopefully increase the robustness and replicability of our results.

Our error taxonomy and its notation are inspired by \citet{thomsonGoldStandardMethodology2020} and \citet{thomson2023evaluating}. After preliminary examination of the outputs, we settled on four error categories: \errinc{INCORRECT}, \errnc{NOT\_CHECKABLE}, \errmis{MISLEADING}, and \errother{OTHER}. To set clear boundaries between the categories, simplify the annotation interface and reach better inter-annotator agreement, we decided to keep our taxonomy more high-level than \citeauthor{thomsonGoldStandardMethodology2020} and not to distinguish between fine-grained categories (e.g., \textit{incorrect name} vs. \textit{incorrect number}). The descriptions of our error categories, as presented in the instructions for annotation, are included in \autoref{tab:errors}.

\begin{table*}[t]
  \small
  \centering
  \begin{tabular}{p{3cm}p{11.75cm}} \toprule                                                                                          \\
    \textbf{Error}         & \textbf{Description}                                                                        \\ \midrule
    \errinc{INCORRECT}     & The fact in the text contradicts the data.                                                  \\
    \errnc{NOT\_CHECKABLE} & The fact in the text cannot be checked given the data.                                      \\
    \errmis{MISLEADING}    & The fact in the text is misleading in the given context.                                    \\
    \errother{OTHER}       & The text is problematic for another reason, e.g., grammatically or stylistically incorrect,
    irrelevant, or repetitive.                                                                                           \\\midrule

    \textbf{Example}       &                                                                                             \\

    \textit{data}          &
    \textbf{Nokia 3310} |
    \textit{color}: black, blue, grey |
    \textit{display}: 320x240px                                                                                          \\
    \textit{text}          &
    Nokia 3310 is \errnc{produced in Finland} and features a \errinc{320x320} display. It is \errmis{available in black color}. \errother{The data seem to provide only partial information about the phone.}
    \vspace*{0.1cm}
    \\ \bottomrule
  \end{tabular}
  \caption{Categories of errors annotated in our evaluation and an example demonstrating the error types. See Appendix \ref{app:humeval} for an explanation of individual errors in the example.}
  \label{tab:errors}
\end{table*}

\subsection{Human-based Evaluation}
\label{sec:humaneval}


For the human annotation metric, we prepared a custom web interface, where 
an annotator is instructed to annotate text spans with respective error categories. We created custom visualizations for each data format (see Figure~\ref{fig:process} and Appendix~\ref{app:outputs} for examples).\footnote{We open-sourced our annotation framework as a stand-alone software package, see \url{https://github.com/kasnerz/factgenie}.}

We hired annotators on the Prolific\footnote{\url{https://prolific.com}} crowdsourcing platform. In total, we hired 100 annotators, each annotating 20 examples (four model outputs for each of the five domains). We selected annotators with at least 10 completed tasks, a 100\% approval rate and English as their primary language. We paid the annotators \pounds 9 per hour, according to the platform's recommendations. The median time for completing the annotations was 47 minutes. See Appendix \ref{app:humeval} for the instructions for the annotators and the annotation interface.

\subsection{GPT-4-based Evaluation}
\label{sec:gpt4eval}

For automatic evaluation, we leverage the fact that LLM-based metrics can be customized for a particular task without the need for training data. In our experiments, we use a metric based on GPT-4 (\texttt{gpt-4-1106-preview}, \citealp{openai2023gpt4}), which was shown to be superior to other LLMs in following fine-grained instructions, reaching high correlations with human judgment on evaluating generated texts \cite{zhaoInvestigatingTabletoTextGeneration2023,sottanaEvaluationMetricsEra2023,kocmiGEMBAMQMDetectingTranslation2023,kocmiLargeLanguageModels2023}.\footnote{We confirmed that GPT-3.5 and Llama 3 have lower correlations with human judgments also in our scenario, see Appendix~\ref{app:openllmeval}.}

We instantiate \gptmetric{} with a prompt and a system message describing the task. We instruct the model to produce a JSON output with sequentially ordered errors using the following format:

\small
\begin{verbatim}
{
  "errors": [{
    "reason": [REASON],
    "text": [TEXT_SPAN],
    "type": [ERROR_CATEGORY]
    }, 
    ...]
}.
\end{verbatim}
\normalsize

Note that we require that the model first generates the free-form text \textit{reason} for the error. Generating the reason comes at almost no extra cost and our cursory observations suggest that requiring it leads to more precise outputs.\footnote{We did not ask the crowdworkers for free-form reasoning about the errors since that would make the annotation notably more complex.}

We align the model outputs with the original text by string matching on \texttt{TEXT\_SPAN}, moving the current position forward after each match. We ensure that the model response is a valid JSON using OpenAI's \texttt{response\_format} parameter.  See Appendix \ref{app:gpt4eval} for more details about the metric, including the prompt and the system message.



\section{Results and Discussion}
\label{sec:discussion}

A summary of the word-level annotations is in \autoref{tab:results_agg} and \ref{tab:results_errperex}, with detailed results per domain provided in Appendix \ref{app:full_results}.

\begin{table*}[ht]
  \small
  \centering
  \begin{tabular}{lccccccccccr}
    \toprule
                     & \multicolumn{2}{c}{\textbf{Incorrect}} & \multicolumn{2}{c}{\textbf{Not Checkable}} & \multicolumn{2}{c}{\textbf{Misleading}} & \multicolumn{2}{c}{\textbf{Other}} & \multicolumn{2}{c}{\textbf{All categories}} &                                                                                                    \\
                     & \humanmetric{}                         & \gptmetric{}                               & \humanmetric{}                          & \gptmetric{}                       & \humanmetric{}                              & \gptmetric{}  & \humanmetric{} & \gptmetric{}  & \humanmetric{} & \gptmetric{}  & \# \textbf{Tok.} \\\midrule
    \textbf{Llama 2} & 1.57                                   & \textbf{2.79}                              & 1.25                                    & 0.91                               & 0.25                                        & \textbf{0.12} & \textbf{0.10}  & 0.09          & 3.18           & 3.90          & 83.8             \\
    \textbf{Mistral} & 2.03                                   & 3.23                                       & 1.12                                    & 0.54                               & 0.44                                        & 0.26          & 0.25           & 0.10          & 3.85           & 4.12          & 114.9            \\
    \textbf{Zephyr}  & \textbf{1.44}                          & 2.84                                       & \textbf{0.77}                           & \textbf{0.40}                      & \textbf{0.20}                               & 0.29          & 0.16           & \textbf{0.05} & \textbf{2.58}  & \textbf{3.58} & 98.0             \\ \cdashlinelr{1-12}
    \textbf{GPT-3.5} & 0.65                                   & 1.76                                       & 0.49                                    & 0.38                               & 0.18                                        & 0.26          & 0.07           & 0.02          & 1.39           & 2.42          & 84.9             \\ \bottomrule
  \end{tabular}
  \caption{The average \textit{numbers of errors per output} (lower is better) based on human annotators (\humanmetric{}) and GPT-4 (\gptmetric{}). We also include the average number of tokens per output in the rightmost column. The results of the best open LLM  are emphasized.}
  \label{tab:results_agg}
\end{table*}

\begin{table*}[ht]
  \small
  \centering
  \begin{tabular}{lrrrrrrrrrr}
    \toprule
                     & \multicolumn{2}{c}{\textbf{Incorrect}} & \multicolumn{2}{c}{\textbf{Not Checkable}} & \multicolumn{2}{c}{\textbf{Misleading}} & \multicolumn{2}{c}{\textbf{Other}} & \multicolumn{2}{c}{\textbf{All categories}}                                                                                        \\
                     & \humanmetric{}                         & \gptmetric{}                               & \humanmetric{}                          & \gptmetric{}                       & \humanmetric{}                              & \gptmetric{}   & \humanmetric{} & \gptmetric{}   & \humanmetric{}  & \gptmetric{}    \\\midrule
    \textbf{Llama 2} & 53.2\%                                 & 80.0\%                                     & 57.4\%                                  & 44.8\%                             & 17.4\%                                      & \textbf{8.8\%} & \textbf{7.6\%} & 7.6\%          & 85.6\%          & 94.0\%          \\
    \textbf{Mistral} & 53.6\%                                 & 80.2\%                                     & 49.6\%                                  & 31.8\%                             & 20.6\%                                      & 17.0\%         & 13.6\%         & 8.4\%          & 81.2\%          & 93.0\%          \\
    \textbf{Zephyr}  & \textbf{46.8\%}                        & \textbf{78.0\%}                            & \textbf{42.2\%}                         & \textbf{25.0\%}                    & \textbf{16.2\%}                             & 20.6\%         & 11.6\%         & \textbf{4.2\%} & \textbf{75.6\%} & \textbf{89.4\%} \\\cdashlinelr{1-11}
    \textbf{GPT-3.5} & 38.0\%                                 & 65.0\%                                     & 28.8\%                                  & 19.6\%                             & 12.6\%                                      & 16.2\%         & 6.2\%          & 2.2\%          & 60.6\%          & 75.8\%          \\\bottomrule
  \end{tabular}
  \caption{The percentage of \textit{outputs containing at least one error} (lower is better) based on human annotators (\humanmetric{}) and  GPT-4 (\gptmetric{}). The results of the best open LLM  are emphasized.}
  \label{tab:results_errperex}
\end{table*}

\subsection{How Accurate Are the Model Outputs?}
Depending on the model, between 76-86\% of examples contain an error according to \humanmetric, suggesting that open LLMs make semantic errors very often. According to \gptmetric, the number is as high as 89-94\%.

The most common error type is \errinc{INCORRECT}. As shown in \autoref{tab:results_agg}, all the open LLMs make more than \textbf{two statements contradicting the data per output on average}. The \errnc{NOT\_CHECKABLE} errors are also relatively common: more than one per output on average according to \humanmetric, and at least one being present in more than 25\% of examples according to both metrics.

The results vary widely according to the domain (see Appendix \ref{app:full_results}). For example, the outputs in \texttt{wikidata} contain much more \errnc{NOT\_CHECKABLE} errors on average (1.01 per output according to \humanmetric) than  \errinc{INCORRECT} errors (0.11 per output according to \humanmetric), suggesting that with simpler inputs, the models tend to introduce extra information. The \texttt{openweather} domain seems to be the most complex with the longest outputs (\textasciitilde 164 tokens), more than eight errors in the output on average, and >90\% of outputs containing an error.

The differences between the open LLMs are not major. Out of the open LLMs, Zephyr has the best results across categories and metrics, followed by Llama 2. However, the outputs of Mistral are longer on average, leaving more space for errors. GPT-3.5 (which we consider separately) does generally better according to both \gptmetric and \humanmetric, although it still makes an error in 60-75\% of examples (2 errors per example on average). In general, the results show that LLMs make too many semantic errors to be usable in practice for D2T generation in a zero-shot setting.

\subsection{Do Evaluation Methods Agree?}
\label{sec:metricscorr}
To quantify the agreement of \humanmetric{} and \gptmetric{}, we computed the Pearson correlation coefficient between the error counts on the level of words, examples, and domains as follows (note that each error category was considered separately):
\begin{itemize}
  \item For $r_{\text{domain}}$, we used the average error counts per domain (see \autoref{tab:results_full}).
  \item For $r_{\text{example}}$, we used the count of errors per example.
  \item For $r_{\text{word}}$, we used the binary indicators marking an error per word.
\end{itemize}
The correlation on the level of words is weak ($r_{\text{word}}=0.26$) but gets better on the example-level ($r_{\text{example}}=0.52$) and even better on the domain-level ($r_{\text{domain}}=0.93$). In \autoref{tab:errors_wordlevel}, we show the percentage of words marked as errors by individual metrics. The metrics agree on the specific words in less than 6\%, although they both mark around 21\% of words as erroneous.

We also measure inter-annotator agreement between human annotators. For that, we obtained annotations from two annotators for 100 model outputs. The results are similar: the annotators agree weakly on the word level ($r_{\text{word}}=0.36$), stronger on the example level ($r_{\text{example}}=0.53$), and even stronger on the domain level ($r_{\text{domain}}=0.85$). We conclude that while the details regarding error spans and categories may vary, the annotators as well as GPT-4 generally agree on the accuracy of model outputs for a given set of examples. In the future, the agreement could be improved by measuring errors on the phrase level \cite{vamvas2022little}.


\begin{table}[t]
  \small
  \centering
  \begin{tabular}{lrrr}
    \toprule
    {}                      & \gptmetric{} & \humanmetric{} & \gptmetric{} + \humanmetric{} \\
    \midrule
    \bfseries Incorrect     & 10.1\%       & 14.2\%         & 4.1\%                         \\
    \bfseries Not checkable & 7.8\%        & 4.3\%          & 2.0\%                         \\
    \bfseries Misleading    & 2.2\%        & 1.5\%          & 0.1\%                         \\
    \bfseries Other         & 1.8\%        & 0.7\%          & 0.1\%                         \\
    \bfseries Total         & 21.9\%       & 20.7\%         & 6.3\%                         \\
    \bottomrule
  \end{tabular}
  \caption{The percentage of \textit{words marked as erroneous} by human annotators (\humanmetric), GPT-4 (\gptmetric), and both approaches at the same time (\humanmetric + \gptmetric).}
  \label{tab:errors_wordlevel}
\end{table}


\subsection{Recommendations for Future Work}
\label{sec:future}
\paragraph{Focus on semantic accuracy.} The output of LLMs is satisfactory regarding the style, format, and purpose of the text. However, the amount of semantic errors remains very high. Improving the semantic accuracy of the models
\cite{li2022faithfulness}, along with new model-based evaluation metrics \cite{liuGEvalNLGEvaluation2023,xuINSTRUCTSCOREExplainableText2023}, could thus help to bring improve LLM-based D2T generation systems where it is most needed.

\paragraph{Use efficient and long-context models.} The memory issues with long context, making few-shot experiments infeasible, can potentially be solved by using more efficient models equipped with Flash Attention \cite{dao2022flashattention} and fast inference libraries such as \texttt{llama.cpp}\footnote{\url{https://github.com/ggerganov/llama.cpp}}. The recent emergence of capable long-context models \cite{bai2023longbench,munkhdalai2024leave} also seems to play in favor of LLM-based D2T generation with long inputs.

\paragraph{Be careful about subtle bugs.} During our preliminary experiments, we fixed several subtle bugs in our API calls such as incorrect instruction templates\footnote{\url{https://huggingface.co/docs/transformers/chat_templating}} or involuntary input truncation. Therefore, we recommend careful checks of API calls, as with the apparent ease of API access and robustness of LLMs, such bugs could go unnoticed and artificially skew the model performance.

\paragraph{Test the models in the wild.} Except for using an ad-hoc dataset of real-world data as we did in our work, the ecological validity of D2T evaluation can also be ensured by continuous evaluation with human users \cite{zheng2023judging} and evaluating the real-world impact of the systems \cite{reiter2023impact}.

\paragraph{Multilinguality is an opportunity.} With the recent efforts in extending D2T generation to low-resource languages \cite{cripwell-etal-2023-2023}, multilingual D2T generation with open LLMs seems a promising direction. Although we did not go beyond English, initial steps were already done by works such as \citet{lorandi2023data}.

\section{Related Work}
\label{sec:relwork}

\subsection{Evaluation of Generated Texts}
Evaluation of generated texts is a complex task lacking a generally accepted solution \cite{celikyilmazEvaluationTextGeneration2021}. While researchers are acknowledging the importance of combining multiple evaluation metrics  \cite{gehrmannGEMBenchmarkNatural2021,gehrmann2022repairing}, most evaluation is still based on comparing the model outputs to human-written references, which tend to be noisy and expensive to collect \cite{dusekSemanticNoiseMatters2019,dusekEvaluatingStateoftheartEndtoEnd2020}.

Many works recently investigated the potential of using LLMs for automatic reference-free evaluation of generated texts, generally achieving high correlations with human judgment \cite{zhaoInvestigatingTabletoTextGeneration2023,sottanaEvaluationMetricsEra2023,kocmiGEMBAMQMDetectingTranslation2023,kocmiLargeLanguageModels2023,chiang-lee-2023-large,wangChatGPTGoodNLG2023a,fu2023gptscore}. However, they also voice concerns about its non-reproducibility \cite{kocmiGEMBAMQMDetectingTranslation2023} and potential bias of these models \cite{wangLargeLanguageModels2023}.


Human evaluation is an essential component of natural language generation experiments \cite{van2019best,vanderleeHumanEvaluationAutomatically2021}. The closest human evaluation protocol to our scenario is the reference-free word-level annotation of complex D2T generation output proposed in \citet{thomsonGoldStandardMethodology2020} and \citet{thomson2023evaluating}.

\subsection{D2T Generation Tasks}

\paragraph{Weather Forecasts} First attempts for generating weather forecasts include template-based and statistical approaches \cite{belz2005corpus,belz2008automatic,angeli-etal-2010-simple} for the Sumtime-meteo and WeatherGov datasets \cite{sripada2002sumtime,liang2009learning}.
More recently, \citet{balakrishnan2019constrained} introduced a weather forecast dataset with tree-structured meaning representations.
Our weather forecasts are less structured and based on a 5-day weather outlook.

\paragraph{Product Descriptions} Our phone specifications are closest to \citet{wen2015toward,wen2016multi}, who introduced a dataset for generating descriptions of laptops and TVs. Their solution was based on recurrent neural networks, although templates remained a go-to approach for the task \cite{wang2017statistical}. Recently, \citet{shaoControllableDiverseText2021} and \citet{kotoCanPretrainedLanguage2022} also proposed specialized architectures based on pretrained language models for the data from big e-commerce platforms.

\paragraph{Sport Reports} All the D2T generation datasets from the Rotowire family \cite{wiseman2017challenges,wangRevisitingChallengesDatatoText2019}, including SportSett:Basketball \cite{thomsonSportSettBasketballRobust2021}, and ESPN-NBA \cite{nieOperationguidedNeuralNetworks2018} focus on generating basketball reports. Along with MLB \cite{puduppullyDatatotextGenerationEntity2019}, these datasets belong among the most challenging D2T datasets, attracting various neural-based solutions \cite{puduppully2019data,puduppully2022data,puduppullyDatatotextGenerationMacro2021,rebuffelHierarchicalModelDatatoText2019}. We use instead simpler data covering ice hockey game summaries.

\paragraph{Chart Captions} Following the early rule-based approaches \cite{demirGeneratingTextualSummaries2008,demirSummarizingInformationGraphics2012}, the approaches for chart captioning recently tackle large-scale datasets from data analytic institutions \cite{obeidCharttoTextGeneratingNatural2020,kantharaj2022chart}. We focus on one of the tasks from \citet{sharmaTCubeDomainAgnosticNeural2021}, which is generating descriptions of time series in the health domain.

\paragraph{Entity Descriptions} The task of generating descriptions for a knowledge graph has been covered extensively in D2T generation \cite[\textit{inter alia}]{gardent2017webnlg,ferreira20202020,agarwal2021knowledge,chen-etal-2020-kgpt,ribeiro2020investigating}. Our task is to describe an entity provided a list of its properties, which is closely related to generating entity descriptions from Wikipedia infotables \cite{lebret2016neural}.

\subsection{D2T Generation with LLMs}
Recent works have focused on exploring the capabilities of closed LLMs on existing D2T generation datasets.
\citet{axelssonUsingLargeLanguage2023} evaluated GPT-3.5 \cite{chatgpt} on WebNLG, along with \citet{yuanEvaluatingGenerativeModels2023}, who also tested the model on the AGENDA dataset \cite{koncel2019text}. Both works found that regardless of potential data contamination, the LLMs rank behind state-of-the-art finetuned models on automatic metrics. \citet{zhaoInvestigatingTabletoTextGeneration2023} tested closed models on modified table-to-text generation datasets and found out that in terms of faithfulness, GPT-4 can outperform state-of-the-art models.


\section{Conclusion}
\label{sec:conclusion}
We provided an exploratory study into D2T generation with open LLMs. We proposed new directions for D2T generation, including using ad-hoc test sets, data in common formats, and reference-free evaluation. By a combination of GPT-4-based metric and human evaluation, we evaluated the performance of LLMs on five domains, providing word-level annotations of model outputs across five domains and recommendations for future directions in D2T generation.

\section*{Acknowledgements}

This work was funded by the European Union (ERC, NG-NLG, 101039303) and Charles University project SVV 260~698. It used resources of the LINDAT/CLARIAH-CZ Research Infrastructure (Czech Ministry of Education, Youth, and Sports project No. LM2018101).

\section*{Limitations}
In our work, we do not include a comparison to other D2T generation approaches. The main reason is that our benchmark is reference-free, while a large majority of prior approaches are based on models finetuned on reference outputs. However, we believe that our work still satisfies our main goal of providing insights into behaviors of open LLM models on D2T generation.

We acknowledge that reference-free metrics currently have various shortcomings, including reliance on closed models and specific human annotation protocols, leading to limited replicability and a high price of execution. The approaches occasionally produce incorrect outcomes themselves and they have only moderate correlations with each other. We believe that these shortcomings will be addressed in the future with open model-based metrics.

Our choice of models is limited to 7B-parameter open LLMs due to our limited computational resources. Also, unlike some other LLMs such as GPT-Neo \cite{black2022gpt} or BLOOM \cite{workshop2022bloom}, the models we used do not disclose the data they were trained on. For this reason, we find it ever more important to test the models on benchmarks whose labels could have \textit{not} been included in their training data.

The approaches based on LLMs may produce factually and semantically incorrect information. Any text produced by the LLMs therefore needs to be carefully examined, and no decisions should be based on the generated text alone. Discovering the \emph{causes} of LLMs' hallucinations is out of scope of this paper,  but is currently a major topic under investigation \cite{ji2023survey}.

\section*{Ethical Considerations}

The human evaluation study was approved by the internal ethics committee of our institution. The annotators were hired over Prolific and paid the platform-recommended wage of 9 GBP/hour.
The annotators were preselected based on their primary language (English) and their country of residence (US, UK, Ireland, Australia, New Zealand).
All annotators were shown detailed instructions and explanation of the data types, data sources, and the purpose of the research (see Appendix~\ref{app:humeval} for details).
The domains in \datatool{} were selected so that they do not contain any sensitive or potentially offensive content.
We do not collect any demographic data about the participants.

\bibliography{custom}

\begin{thebibliography}{89}
\expandafter\ifx\csname natexlab\endcsname\relax\def\natexlab#1{#1}\fi

\bibitem[{Agarwal et~al.(2021)Agarwal, Ge, Shakeri, and Al{-}Rfou}]{agarwal2021knowledge}
Oshin Agarwal, Heming Ge, Siamak Shakeri, and Rami Al{-}Rfou. 2021.
\newblock \href {https://doi.org/10.18653/V1/2021.NAACL-MAIN.278} {{{K}nowledge} {{G}raph} {{B}ased} {{S}ynthetic} {{C}orpus} {{G}eneration} for {{K}nowledge-Enhanced} {{L}anguage} {{M}odel} {{P}re-training}}.
\newblock In \emph{Proceedings of the 2021 Conference of the North American Chapter of the Association for Computational Linguistics: Human Language Technologies, {NAACL-HLT} 2021}, pages 3554--3565, Online.

\bibitem[{Aiyappa et~al.(2023)Aiyappa, An, Kwak, and Ahn}]{aiyappa-etal-2023-trust}
Rachith Aiyappa, Jisun An, Haewoon Kwak, and Yong-yeol Ahn. 2023.
\newblock \href {https://doi.org/10.18653/v1/2023.trustnlp-1.5} {{{C}an} we trust the evaluation on {{C}hat{GPT}?}}
\newblock In \emph{Proceedings of the 3rd Workshop on Trustworthy Natural Language Processing (TrustNLP 2023)}, pages 47--54, Toronto, Canada.

\bibitem[{Angeli et~al.(2010)Angeli, Liang, and Klein}]{angeli-etal-2010-simple}
Gabor Angeli, Percy Liang, and Dan Klein. 2010.
\newblock \href {https://aclanthology.org/D10-1049/} {{A} {{S}imple} {{D}omain-Independent} {{P}robabilistic} {{A}pproach} to {{G}eneration}}.
\newblock In \emph{Proceedings of the 2010 Conference on Empirical Methods in Natural Language Processing, {EMNLP} 2010, 9-11 October 2010, {MIT} Stata Center, Massachusetts, USA, {A} meeting of SIGDAT, a Special Interest Group of the {ACL}}, pages 502--512.

\bibitem[{Axelsson and Skantze(2023)}]{axelssonUsingLargeLanguage2023}
Agnes Axelsson and Gabriel Skantze. 2023.
\newblock \href {https://doi.org/10.48550/ARXIV.2307.07312} {{{U}sing} {{L}arge} {{L}anguage} {{M}odels} for {{Z}ero-Shot} {{N}atural} {{L}anguage} {{G}eneration} from {{K}nowledge} {{G}raphs}}.
\newblock \emph{CoRR}, abs/2307.07312.

\bibitem[{Bai et~al.(2023)Bai, Lv, Zhang, Lyu, Tang, Huang, Du, Liu, Zeng, Hou, Dong, Tang, and Li}]{bai2023longbench}
Yushi Bai, Xin Lv, Jiajie Zhang, Hongchang Lyu, Jiankai Tang, Zhidian Huang, Zhengxiao Du, Xiao Liu, Aohan Zeng, Lei Hou, Yuxiao Dong, Jie Tang, and Juanzi Li. 2023.
\newblock \href {https://doi.org/10.48550/ARXIV.2308.14508} {{LongBench}: {A} bilingual, multitask benchmark for long context understanding}.
\newblock \emph{CoRR}, abs/2308.14508.

\bibitem[{Balakrishnan et~al.(2019)Balakrishnan, Rao, Upasani, White, and Subba}]{balakrishnan2019constrained}
Anusha Balakrishnan, Jinfeng Rao, Kartikeya Upasani, Michael White, and Rajen Subba. 2019.
\newblock \href {https://doi.org/10.18653/V1/P19-1080} {{{C}onstrained} {{D}ecoding} for {{N}eural} {NLG} from {{C}ompositional} {{R}epresentations} in {{T}ask-Oriented} {{D}ialogue}}.
\newblock In \emph{Proceedings of the 57th Conference of the Association for Computational Linguistics, {ACL} 2019, Volume 1: Long Papers}, pages 831--844, Florence, Italy.

\bibitem[{Balloccu et~al.(2024)Balloccu, Schmidtov{\'{a}}, Lango, and Dušek}]{balloccu2024leak}
Simone Balloccu, Patr{\'{\i}}cia Schmidtov{\'{a}}, Mateusz Lango, and Ondrej Dušek. 2024.
\newblock \href {https://aclanthology.org/2024.eacl-long.5} {Leak, cheat, repeat: Data contamination and evaluation malpractices in closed-source {LLMs}}.
\newblock In \emph{Proceedings of the 18th Conference of the European Chapter of the Association for Computational Linguistics, {EACL} 2024 - Volume 1: Long Papers}, pages 67--93, St. Julian's, Malta.

\bibitem[{Beeching et~al.(2023)Beeching, Fourrier, Habib, Han, Lambert, Rajani, Sanseviero, Tunstall, and Wolf}]{open-llm-leaderboard}
Edward Beeching, Clémentine Fourrier, Nathan Habib, Sheon Han, Nathan Lambert, Nazneen Rajani, Omar Sanseviero, Lewis Tunstall, and Thomas Wolf. 2023.
\newblock {{O}pen} {L}{L}{M} {{L}eaderboard}.
\newblock \url{https://huggingface.co/spaces/HuggingFaceH4/open_llm_leaderboard}.

\bibitem[{Belz(2005)}]{belz2005corpus}
Anja Belz. 2005.
\newblock {{C}orpus-driven} generation of weather forecasts.
\newblock In \emph{Proc. 3rd Corpus Linguistics Conference}.

\bibitem[{Belz(2008)}]{belz2008automatic}
Anja Belz. 2008.
\newblock \href {https://doi.org/10.1017/S1351324907004664} {Automatic generation of weather forecast texts using comprehensive probabilistic generation-space models}.
\newblock \emph{Nat. Lang. Eng.}, 14(4):431--455.

\bibitem[{{BigScience Workshop} et~al.(2022){BigScience Workshop}, Scao, Fan, Akiki, Pavlick, Ili{\'c}, Hesslow, Castagn{\'e}, Luccioni, Yvon et~al.}]{workshop2022bloom}
{BigScience Workshop}, Teven~Le Scao, Angela Fan, Christopher Akiki, Ellie Pavlick, Suzana Ili{\'c}, Daniel Hesslow, Roman Castagn{\'e}, Alexandra~Sasha Luccioni, Fran{\c{c}}ois Yvon, et~al. 2022.
\newblock \href {https://arxiv.org/abs/2211.05100} {Bloom: A 176b-parameter open-access multilingual language model}.
\newblock \emph{arXiv preprint arXiv:2211.05100}.

\bibitem[{Black et~al.(2022)Black, Biderman, Hallahan, Anthony, Gao, Golding, He, Leahy, McDonell, Phang et~al.}]{black2022gpt}
Sidney Black, Stella Biderman, Eric Hallahan, Quentin Anthony, Leo Gao, Laurence Golding, Horace He, Connor Leahy, Kyle McDonell, Jason Phang, et~al. 2022.
\newblock {GPT-NeoX-20B}: An open-source autoregressive language model.
\newblock In \emph{Proceedings of BigScience Episode\# 5--Workshop on Challenges \& Perspectives in Creating Large Language Models}, pages 95--136.

\bibitem[{Boschin(2019)}]{boschin2019wikidatasets}
Armand Boschin. 2019.
\newblock \href {http://arxiv.org/abs/1906.04536} {{WikiDataSets} : Standardized sub-graphs from {WikiData}}.
\newblock \emph{CoRR}, abs/1906.04536.

\bibitem[{Castro~Ferreira et~al.(2020)Castro~Ferreira, Gardent, Ilinykh, van~der Lee, Mille, Moussallem, and Shimorina}]{ferreira20202020}
Thiago Castro~Ferreira, Claire Gardent, Nikolai Ilinykh, Chris van~der Lee, Simon Mille, Diego Moussallem, and Anastasia Shimorina. 2020.
\newblock \href {https://aclanthology.org/2020.webnlg-1.7} {The 2020 bilingual, bi-directional {W}eb{NLG}+ shared task: Overview and evaluation results ({W}eb{NLG}+ 2020)}.
\newblock In \emph{Proceedings of the 3rd International Workshop on Natural Language Generation from the Semantic Web (WebNLG+)}, pages 55--76, Dublin, Ireland (Virtual). Association for Computational Linguistics.

\bibitem[{Celikyilmaz et~al.(2020)Celikyilmaz, Clark, and Gao}]{celikyilmazEvaluationTextGeneration2021}
Asli Celikyilmaz, Elizabeth Clark, and Jianfeng Gao. 2020.
\newblock \href {http://arxiv.org/abs/2006.14799} {Evaluation of text generation: {A} survey}.
\newblock \emph{CoRR}, abs/2006.14799.

\bibitem[{Chen et~al.(2023)Chen, Zaharia, and Zou}]{chen2023chatgpt}
Lingjiao Chen, Matei Zaharia, and James Zou. 2023.
\newblock \href {https://doi.org/10.48550/ARXIV.2307.09009} {{{H}ow} is {{C}hat{G}{P}{T}'s} behavior changing over time?}
\newblock \emph{CoRR}, abs/2307.09009.

\bibitem[{Chen et~al.(2020)Chen, Su, Yan, and Wang}]{chen-etal-2020-kgpt}
Wenhu Chen, Yu~Su, Xifeng Yan, and William~Yang Wang. 2020.
\newblock \href {https://doi.org/10.18653/V1/2020.EMNLP-MAIN.697} {{KGPT:} {{K}nowledge-Grounded} {{P}re-Training} for {{D}ata-to-Text} {{G}eneration}}.
\newblock In \emph{Proceedings of the 2020 Conference on Empirical Methods in Natural Language Processing, {EMNLP} 2020}, pages 8635--8648, Online.

\bibitem[{Chiang and Lee(2023)}]{chiang-lee-2023-large}
David~Cheng{-}Han Chiang and Hung{-}yi Lee. 2023.
\newblock \href {https://doi.org/10.18653/V1/2023.ACL-LONG.870} {{{C}an} {{L}arge} {{L}anguage} {{M}odels} {{B}e} an {{A}lternative} to {{H}uman} {{E}valuations?}}
\newblock In \emph{Proceedings of the 61st Annual Meeting of the Association for Computational Linguistics (Volume 1: Long Papers), {ACL} 2023}, pages 15607--15631, Toronto, Canada.

\bibitem[{Cripwell et~al.(2023)Cripwell, Belz, Gardent, Gatt, Borg, Borg, Judge, Lorandi, Nikiforovskaya, and Soto~Martinez}]{cripwell-etal-2023-2023}
Liam Cripwell, Anya Belz, Claire Gardent, Albert Gatt, Claudia Borg, Marthese Borg, John Judge, Michela Lorandi, Anna Nikiforovskaya, and William Soto~Martinez. 2023.
\newblock \href {https://aclanthology.org/2023.mmnlg-1.6} {{{T}he} 2023 {W}eb{NLG} {{S}hared} {{T}ask} on {{L}ow} {{R}esource} {{L}anguages.} {{O}verview} and {{E}valuation} {{R}esults} {({W}eb{NLG}} 2023)}.
\newblock In \emph{Proceedings of the Workshop on Multimodal, Multilingual Natural Language Generation and Multilingual WebNLG Challenge (MM-NLG 2023)}, pages 55--66, Prague, Czech Republic.

\bibitem[{Dao et~al.(2022)Dao, Fu, Ermon, Rudra, and R{\'{e}}}]{dao2022flashattention}
Tri Dao, Daniel~Y. Fu, Stefano Ermon, Atri Rudra, and Christopher R{\'{e}}. 2022.
\newblock \href {http://papers.nips.cc/paper\_files/paper/2022/hash/67d57c32e20fd0a7a302cb81d36e40d5-Abstract-Conference.html} {{{F}lash{A}ttention}: {{F}ast} and {{M}emory-Efficient} {{E}xact} {{A}ttention} with {{I}{O}-Awareness}}.
\newblock In \emph{Advances in Neural Information Processing Systems 35: Annual Conference on Neural Information Processing Systems 2022, NeurIPS 2022}, New Orleans, LA, USA.

\bibitem[{Demir et~al.(2008)Demir, Carberry, and McCoy}]{demirGeneratingTextualSummaries2008}
Seniz Demir, Sandra Carberry, and Kathleen~F. McCoy. 2008.
\newblock \href {https://aclanthology.org/W08-1103/} {{{G}enerating} {{T}extual} {{S}ummaries} of {{B}ar} {{C}harts}}.
\newblock In \emph{{INLG} 2008 - Proceedings of the Fifth International Natural Language Generation Conference, June 12-14, 2008, Salt Fork}, Ohio, USA.

\bibitem[{Demir et~al.(2012)Demir, Carberry, and McCoy}]{demirSummarizingInformationGraphics2012}
Seniz Demir, Sandra Carberry, and Kathleen~F. McCoy. 2012.
\newblock \href {https://doi.org/10.1162/COLI\_A\_00091} {{{S}ummarizing} {{I}nformation} {{G}raphics} {{T}extually}}.
\newblock \emph{Comput. Linguistics}, 38(3):527--574.

\bibitem[{Dušek et~al.(2019)Dušek, Howcroft, and Rieser}]{dusekSemanticNoiseMatters2019}
Ondrej Dušek, David~M. Howcroft, and Verena Rieser. 2019.
\newblock \href {https://doi.org/10.18653/V1/W19-8652} {Semantic noise matters for neural natural language generation}.
\newblock In \emph{Proceedings of the 12th International Conference on Natural Language Generation, {INLG} 2019}, pages 421--426, Tokyo, Japan.

\bibitem[{Dušek et~al.(2020)Dušek, Novikova, and Rieser}]{dusekEvaluatingStateoftheartEndtoEnd2020}
Ondrej Dušek, Jekaterina Novikova, and Verena Rieser. 2020.
\newblock \href {https://doi.org/10.1016/J.CSL.2019.06.009} {Evaluating the state-of-the-art of end-to-end natural language generation: The {E2E} {NLG} challenge}.
\newblock \emph{Comput. Speech Lang.}, 59:123--156.

\bibitem[{Fu et~al.(2023)Fu, Ng, Jiang, and Liu}]{fu2023gptscore}
Jinlan Fu, See{-}Kiong Ng, Zhengbao Jiang, and Pengfei Liu. 2023.
\newblock \href {https://doi.org/10.48550/ARXIV.2302.04166} {{{G}{P}{T}{S}core}: {{E}valuate} as {{Y}ou} {{D}esire}}.
\newblock \emph{CoRR}, abs/2302.04166.

\bibitem[{Gardent et~al.(2017)Gardent, Shimorina, Narayan, and Perez{-}Beltrachini}]{gardent2017webnlg}
Claire Gardent, Anastasia Shimorina, Shashi Narayan, and Laura Perez{-}Beltrachini. 2017.
\newblock \href {https://doi.org/10.18653/V1/W17-3518} {{{T}he} {W}eb{N}{L}{G} {{C}hallenge}: {{G}enerating} {{T}ext} from {RDF} {{D}ata}}.
\newblock In \emph{Proceedings of the 10th International Conference on Natural Language Generation, {INLG} 2017, Santiago de Compostela}, pages 124--133, Spain.

\bibitem[{Gatt and Krahmer(2018)}]{gatt2018survey}
Albert Gatt and Emiel Krahmer. 2018.
\newblock \href {https://doi.org/10.1613/JAIR.5477} {{{S}urvey} of the {{S}tate} of the {{A}rt} in {{N}atural} {{L}anguage} {{G}eneration}: {{C}ore} tasks, applications and evaluation}.
\newblock \emph{J. Artif. Intell. Res.}, 61:65--170.

\bibitem[{Gehrmann et~al.(2021)Gehrmann, Adewumi, Aggarwal, Ammanamanchi, Anuoluwapo, Bosselut, Chandu, Clinciu, Das, Dhole, Du, Durmus, Dušek, Emezue, Gangal, Garbacea, Hashimoto, Hou, Jernite, Jhamtani, Ji, Jolly, Kumar, Ladhak, Madaan, Maddela, Mahajan, Mahamood, Majumder, Martins, McMillan{-}Major, Mille, van Miltenburg, Nadeem, Narayan, Nikolaev, Niyongabo, Osei, Parikh, Perez{-}Beltrachini, Rao, Raunak, Rodriguez, Santhanam, Sedoc, Sellam, Shaikh, Shimorina, Cabezudo, Strobelt, Subramani, Xu, Yang, Yerukola, and Zhou}]{gehrmannGEMBenchmarkNatural2021}
Sebastian Gehrmann, Tosin~P. Adewumi, Karmanya Aggarwal, Pawan~Sasanka Ammanamanchi, Aremu Anuoluwapo, Antoine Bosselut, Khyathi~Raghavi Chandu, Miruna{-}Adriana Clinciu, Dipanjan Das, Kaustubh~D. Dhole, Wanyu Du, Esin Durmus, Ondrej Dušek, Chris Emezue, Varun Gangal, Cristina Garbacea, Tatsunori Hashimoto, Yufang Hou, Yacine Jernite, Harsh Jhamtani, Yangfeng Ji, Shailza Jolly, Dhruv Kumar, Faisal Ladhak, Aman Madaan, Mounica Maddela, Khyati Mahajan, Saad Mahamood, Bodhisattwa~Prasad Majumder, Pedro~Henrique Martins, Angelina McMillan{-}Major, Simon Mille, Emiel van Miltenburg, Moin Nadeem, Shashi Narayan, Vitaly Nikolaev, Rubungo~Andre Niyongabo, Salomey Osei, Ankur~P. Parikh, Laura Perez{-}Beltrachini, Niranjan~Ramesh Rao, Vikas Raunak, Juan~Diego Rodriguez, Sashank Santhanam, Jo{\~{a}}o Sedoc, Thibault Sellam, Samira Shaikh, Anastasia Shimorina, Marco Antonio~Sobrevilla Cabezudo, Hendrik Strobelt, Nishant Subramani, Wei Xu, Diyi Yang, Akhila Yerukola, and Jiawei Zhou. 2021.
\newblock \href {http://arxiv.org/abs/2102.01672} {The {GEM} benchmark: Natural language generation, its evaluation and metrics}.
\newblock \emph{CoRR}, abs/2102.01672.

\bibitem[{Gehrmann et~al.(2023)Gehrmann, Clark, and Sellam}]{gehrmann2022repairing}
Sebastian Gehrmann, Elizabeth Clark, and Thibault Sellam. 2023.
\newblock \href {https://doi.org/10.1613/JAIR.1.13715} {{{R}epairing} the {{C}racked} {{F}oundation}: {A} {{S}urvey} of {{O}bstacles} in {{E}valuation} {{P}ractices} for {{G}enerated} {{T}ext}}.
\newblock \emph{J. Artif. Intell. Res.}, 77:103--166.

\bibitem[{Golchin and Surdeanu(2023)}]{golchin2023time}
Shahriar Golchin and Mihai Surdeanu. 2023.
\newblock \href {https://doi.org/10.48550/ARXIV.2308.08493} {{{T}ime} {{T}ravel} in {{L}{L}{M}s}: {{T}racing} {{D}ata} {{C}ontamination} in {{L}arge} {{L}anguage} {{M}odels}}.
\newblock \emph{CoRR}, abs/2308.08493.

\bibitem[{Holtzman et~al.(2023)Holtzman, West, and Zettlemoyer}]{holtzmanGenerativeModelsComplex2023}
Ari Holtzman, Peter West, and Luke Zettlemoyer. 2023.
\newblock \href {https://doi.org/10.48550/ARXIV.2308.00189} {{{G}enerative} {{M}odels} as a {{C}omplex} {{S}ystems} {{S}cience}: {{H}ow} can we make sense of large language model behavior?}
\newblock \emph{CoRR}, abs/2308.00189.

\bibitem[{Ji et~al.(2023)Ji, Lee, Frieske, Yu, Su, Xu, Ishii, Bang, Madotto, and Fung}]{ji2023survey}
Ziwei Ji, Nayeon Lee, Rita Frieske, Tiezheng Yu, Dan Su, Yan Xu, Etsuko Ishii, Yejin Bang, Andrea Madotto, and Pascale Fung. 2023.
\newblock \href {https://doi.org/10.1145/3571730} {Survey of hallucination in natural language generation}.
\newblock \emph{{ACM} Comput. Surv.}, 55(12):248:1--248:38.

\bibitem[{Jiang et~al.(2023)Jiang, Sablayrolles, Mensch, Bamford, Chaplot, de~Las~Casas, Bressand, Lengyel, Lample, Saulnier, Lavaud, Lachaux, Stock, Scao, Lavril, Wang, Lacroix, and Sayed}]{jiangMistral7B2023}
Albert~Q. Jiang, Alexandre Sablayrolles, Arthur Mensch, Chris Bamford, Devendra~Singh Chaplot, Diego de~Las~Casas, Florian Bressand, Gianna Lengyel, Guillaume Lample, Lucile Saulnier, L{\'{e}}lio~Renard Lavaud, Marie{-}Anne Lachaux, Pierre Stock, Teven~Le Scao, Thibaut Lavril, Thomas Wang, Timoth{\'{e}}e Lacroix, and William~El Sayed. 2023.
\newblock \href {https://doi.org/10.48550/ARXIV.2310.06825} {{{M}istral} {7{B}}}.
\newblock \emph{CoRR}, abs/2310.06825.

\bibitem[{Kantharaj et~al.(2022)Kantharaj, Leong, Lin, Masry, Thakkar, Hoque, and Joty}]{kantharaj2022chart}
Shankar Kantharaj, Rixie Tiffany~Ko Leong, Xiang Lin, Ahmed Masry, Megh Thakkar, Enamul Hoque, and Shafiq~R. Joty. 2022.
\newblock \href {https://doi.org/10.18653/V1/2022.ACL-LONG.277} {{{C}hart-to-{T}ext}: {A} {{L}arge-Scale} {{B}enchmark} for {{C}hart} {{S}ummarization}}.
\newblock In \emph{Proceedings of the 60th Annual Meeting of the Association for Computational Linguistics (Volume 1: Long Papers), {ACL} 2022}, pages 4005--4023, Dublin, Ireland.

\bibitem[{Kasner et~al.(2023)Kasner, Konstas, and Dušek}]{kasner2023mind}
Zdeněk Kasner, Ioannis Konstas, and Ondřej Dušek. 2023.
\newblock \href {https://doi.org/10.18653/V1/2023.EACL-MAIN.176} {{{M}ind} the {{L}abels}: {{D}escribing} {{R}elations} in {{K}nowledge} {{G}raphs} {{W}ith} {{P}retrained} {{M}odels}}.
\newblock In \emph{Proceedings of the 17th Conference of the European Chapter of the Association for Computational Linguistics, {EACL} 2023, Dubrovnik}, pages 2390--2407, Croatia.

\bibitem[{Kocmi and Federmann(2023{\natexlab{a}})}]{kocmiGEMBAMQMDetectingTranslation2023}
Tom Kocmi and Christian Federmann. 2023{\natexlab{a}}.
\newblock \href {https://aclanthology.org/2023.wmt-1.64} {{GEMBA-MQM:} {{D}etecting} {{T}ranslation} {{Q}uality} {{E}rror} {{S}pans} with {GPT-4}}.
\newblock In \emph{Proceedings of the Eighth Conference on Machine Translation, {WMT} 2023}, pages 768--775, Singapore.

\bibitem[{Kocmi and Federmann(2023{\natexlab{b}})}]{kocmiLargeLanguageModels2023}
Tom Kocmi and Christian Federmann. 2023{\natexlab{b}}.
\newblock \href {https://aclanthology.org/2023.eamt-1.19} {{{L}arge} {{L}anguage} {{M}odels} {{A}re} {{S}tate-of-the-Art} {{E}valuators} of {{T}ranslation} {{Q}uality}}.
\newblock In \emph{Proceedings of the 24th Annual Conference of the European Association for Machine Translation, {EAMT} 2023}, pages 193--203, Tampere, Finland.

\bibitem[{Koncel{-}Kedziorski et~al.(2019)Koncel{-}Kedziorski, Bekal, Luan, Lapata, and Hajishirzi}]{koncel2019text}
Rik Koncel{-}Kedziorski, Dhanush Bekal, Yi~Luan, Mirella Lapata, and Hannaneh Hajishirzi. 2019.
\newblock \href {https://doi.org/10.18653/V1/N19-1238} {{{T}ext} {{G}eneration} from {{K}nowledge} {{G}raphs} with {{G}raph} {{T}ransformers}}.
\newblock In \emph{Proceedings of the 2019 Conference of the North American Chapter of the Association for Computational Linguistics: Human Language Technologies, {NAACL-HLT} 2019, Minneapolis, MN, Volume 1 (Long and Short Papers)}, pages 2284--2293, USA.

\bibitem[{Koo et~al.(2023)Koo, Lee, Raheja, Park, Kim, and Kang}]{kooBenchmarkingCognitiveBiases2023}
Ryan Koo, Minhwa Lee, Vipul Raheja, Jong~Inn Park, Zae~Myung Kim, and Dongyeop Kang. 2023.
\newblock \href {https://doi.org/10.48550/ARXIV.2309.17012} {Benchmarking cognitive biases in large language models as evaluators}.
\newblock \emph{CoRR}, abs/2309.17012.

\bibitem[{Koto et~al.(2022)Koto, Lau, and Baldwin}]{kotoCanPretrainedLanguage2022}
Fajri Koto, Jey~Han Lau, and Timothy Baldwin. 2022.
\newblock \href {https://doi.org/10.18653/v1/2022.ecnlp-1.27} {{{C}an} {{{Pretrained} {{L}anguage} {{M}odels} {{G}enerate} {{P}ersuasive}},} {{{Faithful}},} and {{{Informative} {{A}d} {T}ext}} for {{{Product} {{D}escriptions}}?}}
\newblock In \emph{Proceedings of the {{Fifth Workshop}} on E-{{Commerce}} and {{NLP}} ({{ECNLP}} 5)}, pages 234--243, {Dublin, Ireland}.

\bibitem[{Lebret et~al.(2016)Lebret, Grangier, and Auli}]{lebret2016neural}
R{\'{e}}mi Lebret, David Grangier, and Michael Auli. 2016.
\newblock \href {https://doi.org/10.18653/V1/D16-1128} {{{N}eural} {{T}ext} {{G}eneration} from {{S}tructured} {{D}ata} with {{A}pplication} to the {{B}iography} {{D}omain}}.
\newblock In \emph{Proceedings of the 2016 Conference on Empirical Methods in Natural Language Processing, {EMNLP} 2016}, pages 1203--1213, Austin, Texas, USA.

\bibitem[{Li et~al.(2022)Li, Wu, Chen, Liu, Xiao, and Wu}]{li2022faithfulness}
Wei Li, Wenhao Wu, Moye Chen, Jiachen Liu, Xinyan Xiao, and Hua Wu. 2022.
\newblock \href {https://doi.org/10.48550/ARXIV.2203.05227} {{{F}aithfulness} in {{N}atural} {{L}anguage} {{G}eneration}: {A} {{S}ystematic} {{S}urvey} of {{A}nalysis,} {{E}valuation} and {{O}ptimization} {{M}ethods}}.
\newblock \emph{CoRR}, abs/2203.05227.

\bibitem[{Liang et~al.(2009)Liang, Jordan, and Klein}]{liang2009learning}
Percy Liang, Michael~I. Jordan, and Dan Klein. 2009.
\newblock \href {https://aclanthology.org/P09-1011/} {{{L}earning} {{S}emantic} {{C}orrespondences} with {{L}ess} {{S}upervision}}.
\newblock In \emph{{ACL} 2009, Proceedings of the 47th Annual Meeting of the Association for Computational Linguistics and the 4th International Joint Conference on Natural Language Processing of the AFNLP, 2-7 August 2009}, pages 91--99, Singapore.

\bibitem[{Liu et~al.(2023)Liu, Iter, Xu, Wang, Xu, and Zhu}]{liuGEvalNLGEvaluation2023}
Yang Liu, Dan Iter, Yichong Xu, Shuohang Wang, Ruochen Xu, and Chenguang Zhu. 2023.
\newblock \href {https://aclanthology.org/2023.emnlp-main.153} {{{G}-{E}val}: {NLG} {{E}valuation} using {{G}pt-4} with {{B}etter} {{H}uman} {{A}lignment}}.
\newblock In \emph{Proceedings of the 2023 Conference on Empirical Methods in Natural Language Processing, {EMNLP} 2023}, pages 2511--2522, Singapore.

\bibitem[{Lorandi and Belz(2023)}]{lorandi2023data}
Michela Lorandi and Anja Belz. 2023.
\newblock \href {https://aclanthology.org/2023.mmnlg-1.9/} {{{D}ata-to-text} {{G}eneration} for {{S}everely} {{U}nder-Resourced} {{L}anguages} with {{G}{P}{T}-3.5}: {A} {{B}it} of {{H}elp} {{N}eeded} from {{G}oogle} {{T}ranslate} {({W}eb{N}{L}{G}} 2023)}.
\newblock In \emph{Proceedings of the Workshop on Multimodal, Multilingual Natural Language Generation and Multilingual WebNLG Challenge (MM-NLG 2023)}, pages 80--86.

\bibitem[{Munkhdalai et~al.(2024)Munkhdalai, Faruqui, and Gopal}]{munkhdalai2024leave}
Tsendsuren Munkhdalai, Manaal Faruqui, and Siddharth Gopal. 2024.
\newblock \href {https://doi.org/10.48550/ARXIV.2404.07143} {Leave no context behind: Efficient infinite context transformers with infini-attention}.
\newblock \emph{CoRR}, abs/2404.07143.

\bibitem[{Nie et~al.(2018)Nie, Wang, Yao, Pan, and Lin}]{nieOperationguidedNeuralNetworks2018}
Feng Nie, Jinpeng Wang, Jin{-}Ge Yao, Rong Pan, and Chin{-}Yew Lin. 2018.
\newblock \href {https://doi.org/10.18653/V1/D18-1422} {{{O}peration-guided} {{N}eural} {{N}etworks} for {{H}igh} {{F}idelity} {{D}ata-{T}o-Text} {{G}eneration}}.
\newblock In \emph{Proceedings of the 2018 Conference on Empirical Methods in Natural Language Processing}, pages 3879--3889, Brussels, Belgium.

\bibitem[{Novikova et~al.(2017)Novikova, Dušek, Curry, and Rieser}]{novikovaWhyWeNeed2017}
Jekaterina Novikova, Ondřej Dušek, Amanda~Cercas Curry, and Verena Rieser. 2017.
\newblock \href {https://doi.org/10.18653/V1/D17-1238} {{{W}hy} {{W}e} {{N}eed} {{N}ew} {{E}valuation} {{M}etrics} for {NLG}}.
\newblock In \emph{Proceedings of the 2017 Conference on Empirical Methods in Natural Language Processing, {EMNLP} 2017}, pages 2241--2252, Copenhagen, Denmark.

\bibitem[{Obeid and Hoque(2020)}]{obeidCharttoTextGeneratingNatural2020}
Jason Obeid and Enamul Hoque. 2020.
\newblock \href {https://aclanthology.org/2020.inlg-1.20/} {{{C}hart-to-{T}ext}: {{G}enerating} {{N}atural} {{L}anguage} {{D}escriptions} for {{C}harts} by {{A}dapting} the {{T}ransformer} {{M}odel}}.
\newblock In \emph{Proceedings of the 13th International Conference on Natural Language Generation, {INLG} 2020}, pages 138--147, Dublin, Ireland.

\bibitem[{OpenAI(2023{\natexlab{a}})}]{openai2023gpt4}
OpenAI. 2023{\natexlab{a}}.
\newblock \href {https://doi.org/10.48550/ARXIV.2303.08774} {{GPT-4} {{T}echnical} {{R}eport}}.
\newblock \emph{CoRR}, abs/2303.08774.

\bibitem[{OpenAI(2023{\natexlab{b}})}]{chatgpt}
OpenAI. 2023{\natexlab{b}}.
\newblock {{I}ntroducing} {C}hat{G}{P}{T}.
\newblock \url{https://openai.com/blog/chatgpt}.
\newblock Accessed on January 9, 2024.

\bibitem[{Ouyang et~al.(2022)Ouyang, Wu, Jiang, Almeida, Wainwright, Mishkin, Zhang, Agarwal, Slama, Ray, Schulman, Hilton, Kelton, Miller, Simens, Askell, Welinder, Christiano, Leike, and Lowe}]{Ouyang2022TrainingLM}
Long Ouyang, Jeffrey Wu, Xu~Jiang, Diogo Almeida, Carroll~L. Wainwright, Pamela Mishkin, Chong Zhang, Sandhini Agarwal, Katarina Slama, Alex Ray, John Schulman, Jacob Hilton, Fraser Kelton, Luke Miller, Maddie Simens, Amanda Askell, Peter Welinder, Paul~F. Christiano, Jan Leike, and Ryan Lowe. 2022.
\newblock \href {http://papers.nips.cc/paper\_files/paper/2022/hash/b1efde53be364a73914f58805a001731-Abstract-Conference.html} {Training language models to follow instructions with human feedback}.
\newblock In \emph{Advances in Neural Information Processing Systems 35: Annual Conference on Neural Information Processing Systems 2022, NeurIPS 2022}, New Orleans, LA, USA.

\bibitem[{Puduppully et~al.(2019{\natexlab{a}})Puduppully, Dong, and Lapata}]{puduppully2019data}
Ratish Puduppully, Li~Dong, and Mirella Lapata. 2019{\natexlab{a}}.
\newblock \href {https://doi.org/10.1609/AAAI.V33I01.33016908} {{{D}ata-to-Text} {{G}eneration} with {{C}ontent} {{S}election} and {{P}lanning}}.
\newblock In \emph{The Thirty-Third {AAAI} Conference on Artificial Intelligence, {AAAI} 2019, The Thirty-First Innovative Applications of Artificial Intelligence Conference, {IAAI} 2019, The Ninth {AAAI} Symposium on Educational Advances in Artificial Intelligence, {EAAI} 2019}, pages 6908--6915, Honolulu, Hawaii, USA.

\bibitem[{Puduppully et~al.(2019{\natexlab{b}})Puduppully, Dong, and Lapata}]{puduppullyDatatotextGenerationEntity2019}
Ratish Puduppully, Li~Dong, and Mirella Lapata. 2019{\natexlab{b}}.
\newblock \href {https://doi.org/10.18653/V1/P19-1195} {{{D}ata-to-text} {{G}eneration} with {{E}ntity} {{M}odeling}}.
\newblock In \emph{Proceedings of the 57th Conference of the Association for Computational Linguistics, {ACL} 2019, Volume 1: Long Papers}, pages 2023--2035, Florence, Italy.

\bibitem[{Puduppully et~al.(2022)Puduppully, Fu, and Lapata}]{puduppully2022data}
Ratish Puduppully, Yao Fu, and Mirella Lapata. 2022.
\newblock \href {https://transacl.org/ojs/index.php/tacl/article/view/3577} {{{D}ata-to-text} {{G}eneration} with {{V}ariational} {{S}equential} {{P}lanning}}.
\newblock \emph{Trans. Assoc. Comput. Linguistics}, 10:697--715.

\bibitem[{Puduppully and Lapata(2021)}]{puduppullyDatatotextGenerationMacro2021}
Ratish Puduppully and Mirella Lapata. 2021.
\newblock \href {https://doi.org/10.1162/TACL\_A\_00381} {{{D}ata-to-text} {{G}eneration} with {{M}acro} {{P}lanning}}.
\newblock \emph{Trans. Assoc. Comput. Linguistics}, 9:510--527.

\bibitem[{Rebuffel et~al.(2020)Rebuffel, Soulier, Scoutheeten, and Gallinari}]{rebuffelHierarchicalModelDatatoText2019}
Cl{\'{e}}ment Rebuffel, Laure Soulier, Geoffrey Scoutheeten, and Patrick Gallinari. 2020.
\newblock \href {https://doi.org/10.1007/978-3-030-45439-5\_5} {{A} {{H}ierarchical} {{M}odel} for {{D}ata-to-Text} {{G}eneration}}.
\newblock In \emph{Advances in Information Retrieval - 42nd European Conference on {IR} Research, {ECIR} 2020, Lisbon, Portugal, April 14-17, 2020, Proceedings, Part {I}}, volume 12035 of \emph{Lecture Notes in Computer Science}, pages 65--80.

\bibitem[{Reiter(2023)}]{reiter2023impact}
Ehud Reiter. 2023.
\newblock {{W}e} should evaluate real-world impact!
\newblock \url{https://ehudreiter.com/2023/11/13/evaluate-real-world-impact/}.
\newblock Accessed on January 11, 2024.

\bibitem[{Reiter and Dale(1997)}]{reiter1997building}
Ehud Reiter and Robert Dale. 1997.
\newblock \href {https://doi.org/10.1017/S1351324997001502} {Building applied natural language generation systems}.
\newblock \emph{Nat. Lang. Eng.}, 3(1):57--87.

\bibitem[{Ribeiro et~al.(2020)Ribeiro, Schmitt, Sch{\"{u}}tze, and Gurevych}]{ribeiro2020investigating}
Leonardo F.~R. Ribeiro, Martin Schmitt, Hinrich Sch{\"{u}}tze, and Iryna Gurevych. 2020.
\newblock \href {http://arxiv.org/abs/2007.08426} {{{I}nvestigating} {{P}retrained} {{L}anguage} {{M}odels} for {{G}raph-to-Text} {{G}eneration}}.
\newblock \emph{CoRR}, abs/2007.08426.

\bibitem[{Rogers(2023)}]{rogers2023closed}
Anna Rogers. 2023.
\newblock {{C}losed} {A}{I} {{M}odels} {{M}ake} {{B}ad} {{B}aselines}.
\newblock \url{https://hackingsemantics.xyz/2023/closed-baselines/}.
\newblock Accessed on January 11, 2024.

\bibitem[{Shao et~al.(2021)Shao, Wang, Lin, Zhang, Zhang, Ji, and Abdelzaher}]{shaoControllableDiverseText2021}
Huajie Shao, Jun Wang, Haohong Lin, Xuezhou Zhang, Aston Zhang, Heng Ji, and Tarek~F. Abdelzaher. 2021.
\newblock \href {https://doi.org/10.1145/3442381.3449838} {{{C}ontrollable} and {{D}iverse} {{T}ext} {{G}eneration} in {{E}-commerce}}.
\newblock In \emph{{WWW} '21: The Web Conference 2021}, pages 2392--2401, Virtual Event / Ljubljana, Slovenia.

\bibitem[{Sharma et~al.(2021)Sharma, Brownstein, and Ramakrishnan}]{sharmaTCubeDomainAgnosticNeural2021}
Mandar Sharma, John~S. Brownstein, and Naren Ramakrishnan. 2021.
\newblock \href {http://arxiv.org/abs/2110.05633} {{{T}{C}ube}: {{D}omain-Agnostic} {{N}eural} {{T}ime-series} {{N}arration}}.
\newblock \emph{CoRR}, abs/2110.05633.

\bibitem[{Sottana et~al.(2023)Sottana, Liang, Zou, and Yuan}]{sottanaEvaluationMetricsEra2023}
Andrea Sottana, Bin Liang, Kai Zou, and Zheng Yuan. 2023.
\newblock \href {https://aclanthology.org/2023.emnlp-main.543} {{{E}valuation} {{M}etrics} in the {{E}ra} of {GPT-4:} {{R}eliably} {{E}valuating} {{L}arge} {{L}anguage} {{M}odels} on {{S}equence} to {{S}equence} {{T}asks}}.
\newblock In \emph{Proceedings of the 2023 Conference on Empirical Methods in Natural Language Processing, {EMNLP} 2023}, pages 8776--8788, Singapore.

\bibitem[{Sripada et~al.(2002)Sripada, Reiter, Hunter, and Yu}]{sripada2002sumtime}
Somayajulu Sripada, Ehud Reiter, Jim Hunter, and Jin Yu. 2002.
\newblock {{S}umtime-meteo}: {{P}arallel} corpus of naturally occurring forecast texts and weather data.
\newblock \emph{Computing Science Department, University of Aberdeen, Aberdeen, Scotland, Tech. Rep. AUCS/TR0201}.

\bibitem[{Stureborg et~al.(2024)Stureborg, Alikaniotis, and Suhara}]{stureborgLargeLanguageModels2024}
Rickard Stureborg, Dimitris Alikaniotis, and Yoshi Suhara. 2024.
\newblock \href {http://arxiv.org/abs/2405.01724} {Large {{{Language} Models}} are {{Inconsistent}} and {{{Biased} Evaluators}}}.

\bibitem[{Thomson and Reiter(2020)}]{thomsonGoldStandardMethodology2020}
Craig Thomson and Ehud Reiter. 2020.
\newblock \href {https://aclanthology.org/2020.inlg-1.22/} {{A} {{G}old} {{S}tandard} {{M}ethodology} for {{E}valuating} {{A}ccuracy} in {{D}ata-{T}o-Text} {{S}ystems}}.
\newblock In \emph{Proceedings of the 13th International Conference on Natural Language Generation, {INLG} 2020}, pages 158--168, Dublin, Ireland.

\bibitem[{Thomson et~al.(2021)Thomson, Reiter, and Sripada}]{thomsonSportSettBasketballRobust2021}
Craig Thomson, Ehud Reiter, and Somayajulu Sripada. 2021.
\newblock {{SportSett}}:{{Basketball}} - {{A}} {{R}obust} and {{M}aintainable} {{D}ataset} for {{{Natural} {{L}anguage} {G}eneration}}.
\newblock page~9.

\bibitem[{Thomson et~al.(2023)Thomson, Reiter, and Sundararajan}]{thomson2023evaluating}
Craig Thomson, Ehud Reiter, and Barkavi Sundararajan. 2023.
\newblock \href {https://doi.org/10.1016/J.CSL.2023.101482} {Evaluating factual accuracy in complex data-to-text}.
\newblock \emph{Comput. Speech Lang.}, 80:101482.

\bibitem[{TogetherAI(2023)}]{llama-2-7b-32k}
TogetherAI. 2023.
\newblock {{P}reparing} for the era of {32{K}} context: {{E}arly} learnings and explorations.
\newblock \url{https://www.together.ai/blog/llama-2-7b-32k}.
\newblock Accessed on January 2, 2024.

\bibitem[{Touvron et~al.(2023{\natexlab{a}})Touvron, Lavril, Izacard, Martinet, Lachaux, Lacroix, Rozi{\`{e}}re, Goyal, Hambro, Azhar, Rodriguez, Joulin, Grave, and Lample}]{touvron2023llama}
Hugo Touvron, Thibaut Lavril, Gautier Izacard, Xavier Martinet, Marie{-}Anne Lachaux, Timoth{\'{e}}e Lacroix, Baptiste Rozi{\`{e}}re, Naman Goyal, Eric Hambro, Faisal Azhar, Aur{\'{e}}lien Rodriguez, Armand Joulin, Edouard Grave, and Guillaume Lample. 2023{\natexlab{a}}.
\newblock \href {https://doi.org/10.48550/ARXIV.2302.13971} {{{L}{L}a{M}{A}}: {{O}pen} and {{E}fficient} {{F}oundation} {{L}anguage} {{M}odels}}.
\newblock \emph{CoRR}, abs/2302.13971.

\bibitem[{Touvron et~al.(2023{\natexlab{b}})Touvron, Martin, Stone, Albert, Almahairi, Babaei, Bashlykov, Batra, Bhargava, Bhosale, Bikel, Blecher, Canton{-}Ferrer, Chen, Cucurull, Esiobu, Fernandes, Fu, Fu, Fuller, Gao, Goswami, Goyal, Hartshorn, Hosseini, Hou, Inan, Kardas, Kerkez, Khabsa, Kloumann, Korenev, Koura, Lachaux, Lavril, Lee, Liskovich, Lu, Mao, Martinet, Mihaylov, Mishra, Molybog, Nie, Poulton, Reizenstein, Rungta, Saladi, Schelten, Silva, Smith, Subramanian, Tan, Tang, Taylor, Williams, Kuan, Xu, Yan, Zarov, Zhang, Fan, Kambadur, Narang, Rodriguez, Stojnic, Edunov, and Scialom}]{touvronLlamaOpenFoundation2023}
Hugo Touvron, Louis Martin, Kevin Stone, Peter Albert, Amjad Almahairi, Yasmine Babaei, Nikolay Bashlykov, Soumya Batra, Prajjwal Bhargava, Shruti Bhosale, Dan Bikel, Lukas Blecher, Cristian Canton{-}Ferrer, Moya Chen, Guillem Cucurull, David Esiobu, Jude Fernandes, Jeremy Fu, Wenyin Fu, Brian Fuller, Cynthia Gao, Vedanuj Goswami, Naman Goyal, Anthony Hartshorn, Saghar Hosseini, Rui Hou, Hakan Inan, Marcin Kardas, Viktor Kerkez, Madian Khabsa, Isabel Kloumann, Artem Korenev, Punit~Singh Koura, Marie{-}Anne Lachaux, Thibaut Lavril, Jenya Lee, Diana Liskovich, Yinghai Lu, Yuning Mao, Xavier Martinet, Todor Mihaylov, Pushkar Mishra, Igor Molybog, Yixin Nie, Andrew Poulton, Jeremy Reizenstein, Rashi Rungta, Kalyan Saladi, Alan Schelten, Ruan Silva, Eric~Michael Smith, Ranjan Subramanian, Xiaoqing~Ellen Tan, Binh Tang, Ross Taylor, Adina Williams, Jian~Xiang Kuan, Puxin Xu, Zheng Yan, Iliyan Zarov, Yuchen Zhang, Angela Fan, Melanie Kambadur, Sharan Narang, Aur{\'{e}}lien Rodriguez, Robert Stojnic, Sergey Edunov,
  and Thomas Scialom. 2023{\natexlab{b}}.
\newblock \href {https://doi.org/10.48550/ARXIV.2307.09288} {{{L}lama} 2: {{O}pen} {{F}oundation} and {{F}ine-Tuned} {{C}hat} {{M}odels}}.
\newblock \emph{CoRR}, abs/2307.09288.

\bibitem[{Tunstall et~al.(2023)Tunstall, Beeching, Lambert, Rajani, Rasul, Belkada, Huang, von Werra, Fourrier, Habib, Sarrazin, Sanseviero, Rush, and Wolf}]{tunstallZephyrDirectDistillation2023}
Lewis Tunstall, Edward Beeching, Nathan Lambert, Nazneen Rajani, Kashif Rasul, Younes Belkada, Shengyi Huang, Leandro von Werra, Cl{\'{e}}mentine Fourrier, Nathan Habib, Nathan Sarrazin, Omar Sanseviero, Alexander~M. Rush, and Thomas Wolf. 2023.
\newblock \href {https://doi.org/10.48550/ARXIV.2310.16944} {{{Z}ephyr}: {{D}irect} {{D}istillation} of {LM} {{A}lignment}}.
\newblock \emph{CoRR}, abs/2310.16944.

\bibitem[{Vamvas and Sennrich(2022)}]{vamvas2022little}
Jannis Vamvas and Rico Sennrich. 2022.
\newblock \href {https://doi.org/10.18653/V1/2022.ACL-SHORT.53} {As little as possible, as much as necessary: Detecting over- and undertranslations with contrastive conditioning}.
\newblock In \emph{Proceedings of the 60th Annual Meeting of the Association for Computational Linguistics (Volume 2: Short Papers), {ACL} 2022}, pages 490--500, Dublin, Ireland.

\bibitem[{van~der Lee et~al.(2021)van~der Lee, Gatt, van Miltenburg, and Krahmer}]{vanderleeHumanEvaluationAutomatically2021}
Chris van~der Lee, Albert Gatt, Emiel van Miltenburg, and Emiel Krahmer. 2021.
\newblock \href {https://doi.org/10.1016/J.CSL.2020.101151} {{{H}uman} evaluation of automatically generated text: {{C}urrent} trends and best practice guidelines}.
\newblock \emph{Comput. Speech Lang.}, 67:101151.

\bibitem[{van~der Lee et~al.(2019)van~der Lee, Gatt, van Miltenburg, Wubben, and Krahmer}]{van2019best}
Chris van~der Lee, Albert Gatt, Emiel van Miltenburg, Sander Wubben, and Emiel Krahmer. 2019.
\newblock \href {https://doi.org/10.18653/V1/W19-8643} {Best practices for the human evaluation of automatically generated text}.
\newblock In \emph{Proceedings of the 12th International Conference on Natural Language Generation, {INLG} 2019}, pages 355--368, Tokyo, Japan.

\bibitem[{Van~Miltenburg et~al.(2023)Van~Miltenburg, Clinciu, Du{\v s}ek, Gkatzia, Inglis, Lepp{\"a}nen, Mahamood, Schoch, Thomson, and Wen}]{vanmiltenburgBarriersEnablingFactors2023}
Emiel Van~Miltenburg, Miruna Clinciu, Ond{\v r}ej Du{\v s}ek, Dimitra Gkatzia, Stephanie Inglis, Leo Lepp{\"a}nen, Saad Mahamood, Stephanie Schoch, Craig Thomson, and Luou Wen. 2023.
\newblock \href {https://doi.org/10.3384/nejlt.2000-1533.2023.4529} {Barriers and enabling factors for error analysis in {{NLG}} research}.
\newblock \emph{Northern European Journal of Language Technology}, 9.

\bibitem[{Wang(2019)}]{wangRevisitingChallengesDatatoText2019}
Hongmin Wang. 2019.
\newblock \href {https://doi.org/10.18653/V1/W19-8639} {{{R}evisiting} {{C}hallenges} in {{D}ata-to-Text} {{G}eneration} with {{F}act} {{G}rounding}}.
\newblock In \emph{Proceedings of the 12th International Conference on Natural Language Generation, {INLG} 2019}, pages 311--322, Tokyo, Japan.

\bibitem[{Wang et~al.(2023{\natexlab{a}})Wang, Liang, Meng, Shi, Li, Xu, Qu, and Zhou}]{wangChatGPTGoodNLG2023a}
Jiaan Wang, Yunlong Liang, Fandong Meng, Haoxiang Shi, Zhixu Li, Jinan Xu, Jianfeng Qu, and Jie Zhou. 2023{\natexlab{a}}.
\newblock \href {https://doi.org/10.48550/ARXIV.2303.04048} {{{I}s} {C}hat{G}{P}{T} a {{G}ood} {NLG} {{E}valuator?} {A} {{P}reliminary} {{S}tudy}}.
\newblock \emph{CoRR}, abs/2303.04048.

\bibitem[{Wang et~al.(2017)Wang, Hou, Liu, Cao, and Lin}]{wang2017statistical}
Jinpeng Wang, Yutai Hou, Jing Liu, Yunbo Cao, and Chin{-}Yew Lin. 2017.
\newblock \href {https://aclanthology.org/I17-2032/} {{A} {{S}tatistical} {{F}ramework} for {{P}roduct} {{D}escription} {{G}eneration}}.
\newblock In \emph{Proceedings of the Eighth International Joint Conference on Natural Language Processing, {IJCNLP} 2017, Volume 2: Short Papers}, pages 187--192, Taipei, Taiwan.

\bibitem[{Wang et~al.(2023{\natexlab{b}})Wang, Li, Chen, Zhu, Lin, Cao, Liu, Liu, and Sui}]{wangLargeLanguageModels2023}
Peiyi Wang, Lei Li, Liang Chen, Dawei Zhu, Binghuai Lin, Yunbo Cao, Qi~Liu, Tianyu Liu, and Zhifang Sui. 2023{\natexlab{b}}.
\newblock \href {https://doi.org/10.48550/ARXIV.2305.17926} {{{L}arge} {{L}anguage} {{M}odels} are not {{F}air} {{E}valuators}}.
\newblock \emph{CoRR}, abs/2305.17926.

\bibitem[{Wen et~al.(2015)Wen, Ga{\v{s}}ic, Mrk{\v{s}}ic, Rojas-Barahona, Su, Vandyke, and Young}]{wen2015toward}
Tsung-Hsien Wen, Milica Ga{\v{s}}ic, Nikola Mrk{\v{s}}ic, Lina~M Rojas-Barahona, Pei-Hao Su, David Vandyke, and Steve Young. 2015.
\newblock \href {https://shawnwun.github.io/papers/slunips2015.pdf} {{{T}oward} multi-domain language generation using recurrent neural networks}.
\newblock In \emph{NIPS Workshop on Machine Learning for Spoken Language Understanding and Interaction}.

\bibitem[{Wen et~al.(2016)Wen, Gasic, Mrksic, Rojas{-}Barahona, Su, Vandyke, and Young}]{wen2016multi}
Tsung{-}Hsien Wen, Milica Gasic, Nikola Mrksic, Lina~Maria Rojas{-}Barahona, Pei{-}Hao Su, David Vandyke, and Steve~J. Young. 2016.
\newblock \href {https://doi.org/10.18653/V1/N16-1015} {{{M}ulti-domain} {{N}eural} {{N}etwork} {{L}anguage} {{G}eneration} for {{S}poken} {{D}ialogue} {{S}ystems}}.
\newblock In \emph{{NAACL} {HLT} 2016, The 2016 Conference of the North American Chapter of the Association for Computational Linguistics: Human Language Technologies}, pages 120--129, San Diego California, USA.

\bibitem[{Wiseman et~al.(2017)Wiseman, Shieber, and Rush}]{wiseman2017challenges}
Sam Wiseman, Stuart~M. Shieber, and Alexander~M. Rush. 2017.
\newblock \href {https://doi.org/10.18653/V1/D17-1239} {{{C}hallenges} in {{D}ata-to-Document} {{G}eneration}}.
\newblock In \emph{Proceedings of the 2017 Conference on Empirical Methods in Natural Language Processing, {EMNLP} 2017}, pages 2253--2263, Copenhagen, Denmark.

\bibitem[{Wolf et~al.(2020)Wolf, Debut, Sanh, Chaumond, Delangue, Moi, Cistac, Rault, Louf, Funtowicz, Davison, Shleifer, von Platen, Ma, Jernite, Plu, Xu, Scao, Gugger, Drame, Lhoest, and Rush}]{wolf2019huggingface}
Thomas Wolf, Lysandre Debut, Victor Sanh, Julien Chaumond, Clement Delangue, Anthony Moi, Pierric Cistac, Tim Rault, R{\'{e}}mi Louf, Morgan Funtowicz, Joe Davison, Sam Shleifer, Patrick von Platen, Clara Ma, Yacine Jernite, Julien Plu, Canwen Xu, Teven~Le Scao, Sylvain Gugger, Mariama Drame, Quentin Lhoest, and Alexander~M. Rush. 2020.
\newblock \href {https://doi.org/10.18653/V1/2020.EMNLP-DEMOS.6} {{{T}ransformers}: {{S}tate-of-the-Art} {{N}atural} {{L}anguage} {{P}rocessing}}.
\newblock In \emph{Proceedings of the 2020 Conference on Empirical Methods in Natural Language Processing: System Demonstrations, {EMNLP} 2020 - Demos}, pages 38--45, Online.

\bibitem[{Xu et~al.(2023)Xu, Wang, Pan, Song, Freitag, Wang, and Li}]{xuINSTRUCTSCOREExplainableText2023}
Wenda Xu, Danqing Wang, Liangming Pan, Zhenqiao Song, Markus Freitag, William~Yang Wang, and Lei Li. 2023.
\newblock \href {http://arxiv.org/abs/2305.14282} {{{{INSTRUCTSCORE}}}: {{{Explainable} {{T}ext} {{G}eneration} {E}valuation}} with {{{Finegrained} {F}eedback}}}.

\bibitem[{Yuan and F{\"{a}}rber(2023)}]{yuanEvaluatingGenerativeModels2023}
Shuzhou Yuan and Michael F{\"{a}}rber. 2023.
\newblock \href {https://aclanthology.org/2023.ranlp-1.133} {{{E}valuating} {{G}enerative} {{M}odels} for {{G}raph-to-Text} {{G}eneration}}.
\newblock In \emph{Proceedings of the 14th International Conference on Recent Advances in Natural Language Processing, {RANLP} 2023}, pages 1256--1264, Varna, Bulgaria.

\bibitem[{Zhao et~al.(2023)Zhao, Zhang, Si, Nan, Tang, and Cohan}]{zhaoInvestigatingTabletoTextGeneration2023}
Yilun Zhao, Haowei Zhang, Shengyun Si, Linyong Nan, Xiangru Tang, and Arman Cohan. 2023.
\newblock \href {http://arxiv.org/abs/2305.14987} {{{I}nvestigating} {{{Table-to-Text} {{G}eneration} {C}apabilities}} of {{LLMs}} in {{{Real-World} {{I}nformation} {{S}eeking} {S}cenarios}}}.

\bibitem[{Zheng et~al.(2023)Zheng, Chiang, Sheng, Zhuang, Wu, Zhuang, Lin, Li, Li, Xing, Zhang, Gonzalez, and Stoica}]{zheng2023judging}
Lianmin Zheng, Wei{-}Lin Chiang, Ying Sheng, Siyuan Zhuang, Zhanghao Wu, Yonghao Zhuang, Zi~Lin, Zhuohan Li, Dacheng Li, Eric~P. Xing, Hao Zhang, Joseph~E. Gonzalez, and Ion Stoica. 2023.
\newblock \href {https://doi.org/10.48550/ARXIV.2306.05685} {{{J}udging} {{L}{L}{M}-as-a-judge} with {{M}{T}-Bench} and {{C}hatbot} {{A}rena}}.
\newblock \emph{CoRR}, abs/2306.05685.

\end{thebibliography}
\bibliographystyle{acl_natbib}

\appendix

\section{\datatool Data}
\label{app:data}

Here, we describe the data sources we include in the \datatool collection tool and the procedure of collecting the \benchmark benchmark. To replicate the data collection, please refer to the scripts we provide.\footnote{\url{https://github.com/kasnerz/quintd}}

\subsection{Selection of Data Sources}
\label{app:dataselection}
When selecting the data sources, we had the following desiderata:

\begin{itemize}
  \item Data needs to be publicly available.
  \item Data needs to represent a common data-to-text task.
  \item Data needs to be in a common format (or straightforwardly transformable to one).
\end{itemize}

We settled on the data sources described in Appendix \ref{app:dataapis}. All the sources can be accessed using an API. Note that some of the APIs have access limits, either for the requests made from a single account per day or for a number of requests from an IP address within a time window. However, these limits do not severely limit the data collection process on the scale we use here.

\subsection{Data Collection}
\autoref{tab:types} summarizes the output types for each domain.

\begin{table}[h]
  \centering
  \begin{tabular}{ll} \toprule                                    \\
    \textbf{Domain Id}   & \textbf{Output type}      \\ \midrule
    \texttt{openweather} & five-day weather forecast \\
    \texttt{gsmarena}    & product description       \\
    \texttt{ice\_hockey} & ice hockey game summary   \\
    \texttt{owid}        & chart caption             \\
    \texttt{wikidata}    & entity description        \\\midrule
  \end{tabular}
  \caption{The output types for individual domains in \datatool.}
  \label{tab:types}
\end{table}

\label{app:dataapis}

\subsubsection{OpenWeather}
\label{app:data_openweather}
OpenWeather (\href{https://openweathermap.org}{OpenWeatherMap.org}) is an online service that provides global weather data via web interface and API. The API responses are in the JSON format \href{https://openweathermap.org/api}{documented} at the official website. For our experiments, we used the \href{https://openweathermap.org/forecast5}{\texttt{forecast5}} API, which allows to download a 5-day forecast with 3-hour resolution for any location specified by its GPS coordinates.

The free tier is limited to 1,000 API calls per day, which is enough to download our whole test set in one bulk. However, at the time of experiments, the free API only allowed to download the data for the \textit{time when the request was made}. At the time of writing, OpenWeather is pushing a new \href{https://openweathermap.org/api/one-call-3}{One Call API 3.0} which allows to download weather data for any timestamp, but only \textit{4 days ahead} (instead of 5). These restrictions somehow limit the replicability of our \benchmark dataset (at least with the free API) but do not limit downloading a new batch of data with a similar format.

For the \benchmark dataset, we randomly sampled 100 cities for each split from the \href{https://public.opendatasoft.com/explore/dataset/geonames-all-cities-with-a-population-1000/table/}{list of cities with a population over 1000} and used their coordinates in the queries to OpenWeather API. All the data forecasts were downloaded on Jan 3, 2024.

\subsubsection{GSMArena}
\label{app:data_gsmarena}
\href{https://www.gsmarena.com}{GSMArena} is a website providing specifications and reviews for mobile devices. For downloading the data, we used the unofficial \href{https://github.com/nordmarin/gsmarena-api}{\texttt{gsmarena-api}} tool, which returns the data in a JSON format. Note that GSMArena imposes limitations on the number of requests per IP address, which may induce delays when downloading a larger amount of data.

To create a balanced sample, we downloaded detailed specifications of 10 products from each available brand and randomly selected 100 products for each split from the downloaded set.

\subsubsection{RapidAPI Ice Hockey}
\label{app:data_icehockey}
\href{https://rapidapi.com}{RapidAPI} is a service that provides API access to data from multiple domains, including sport, finance, entertainment, and others. Most APIs are provided in a freemium mode, i.e., with a limited number of daily API calls.

For \datatool, we selected the \href{https://rapidapi.com/fluis.lacasse/api/icehockeyapi}{\texttt{IceHockeyAPI}} (popularity 9.1 / 10), which provides access to ice hockey games from world top leagues. Our choice was influenced by our own personal preferences, combined with the desire to cover a sport that has not been covered previously in sports report generation.

We used the \href{https://icehockeyapi.p.rapidapi.com/api/ice-hockey/matches}{\texttt{matches}} endpoint which returns high-level details about a game. Note that the API allows only 50 requests per day, but that does not limit the data collection since the endpoint returns \textit{all the games} played on a particular day in a single request. We downloaded the games played on 27 November 2023 for the development set (184 games) and 29 November 2023 for the test set (216 games), taking a random sample of 100 for each split.

\subsubsection{OurWorldInData}
\label{app:data_owid}
\href{https://ourworldindata.org}{OurWorldInData} is a public database and web interface for data about world developments in various domains and sources. We used the official API (currently experimental), which is accessible through the Python package \href{https://pypi.org/project/owid-catalog/}{\texttt{owid-catalog}}. The package allows accessing individual CSV tables as Pandas dataframes.

For our data collection, we decided to limit ourselves to time series, i.e., a single column with values changing over time. Besides the simplicity of visualizing such a chart (which is used by human annotators for checking the correctness of the output), there is also a clear goal for the target chart description: describing the developments of a value over time.
\renewcommand\_{\textunderscore\allowbreak}
We also limited ourselves to the health domain. In particular, we selected the tables  \href{https://ourworldindata.org/coronavirus}{COVID data} (columns
\texttt{new\_cases\_smoothed\_per\_million},
\texttt{new\_tests\_smoothed\_per\_thousand},
\texttt{people\_vaccinated\_per\_hundred},
\texttt{reproduction\_rate}, and
\texttt{positive\_rate}) and \href{https://ourworldindata.org/life-expectancy}{Life expectancy data} (column \texttt{life\_expectancy\_0}).

We downloaded the data for all countries with non-empty entries in the table, taking a random sample of 100 examples for each split. On model input, we formatted the data for each time series as a two-column CSV, including the title, the description, and the unit for each example as a comment (\texttt{\#}) at the beginning of the input.

\subsubsection{Wikidata}
\label{app:data_wikidata}
\href{https://wikidata.org}{Wikidata} is a large open-source knowledge graph containing factual information about entities and their properties. Wikidata provides access through an \href{https://www.wikidata.org/wiki/Wikidata:REST_API}{official API}, but we instead decided to extract our data using the \href{https://graphs.telecom-paris.fr/Home_page.html#wikidatasets-section}{\texttt{wikidatasets}} \cite{boschin2019wikidatasets} Python library, which provides access to preprocessed properties of entities from particular domains. It allowed us to avoid crawling and filtering the knowledge graph, and its offline processing made the data collection faster.\footnote{All the entities and properties are linked with an identifier to the Wikidata database, making the process also replicable through the official API.}

For our dataset, we selected the entities from the \texttt{companies}, \texttt{countries}, \texttt{films}, and \texttt{humans} domains. For each entity, we randomly extracted between 2 to 10 properties in the knowledge graph. We extracted up to 100 subgraphs for each domain and took a random sample of 100 subgraphs for each split. On model input, we formatted each subgraph as a simple Markdown-formatted text snippet, using the entity as a title and including a bullet point for each key-value pair.

\section{Human Evaluation}
\label{app:humeval}

As described in §\ref{sec:humaneval}, we set up the human evaluation campaign on Prolific. To make the data more accessible to the annotators, we created custom data visualizations for each domain. For the data in \texttt{openweather} and \texttt{owid}, we used interactive graphs from \href{https://www.highcharts.com}{Highcharts.com}, and we manually created the tables for other domains. You can find the full instructions for human annotators in \autoref{fig:humaninstructions} and the examples of data visualizations in Appendix \ref{app:outputs}.


\section{GPT-4 Evaluation}
\label{app:gpt4eval}

We used the prompt in \autoref{fig:gpt4prompt} for instantiating the GPT-4-based metric.\footnote{Note that the example in the prompt differs from the example used for human annotators (see \autoref{fig:humaninstructions}). We revised the example to be more instructive, but we were not able to re-run the GPT-4 evaluation due to our limited budget.} We set the temperature to 0 to improve the replicability of the process. We ensured that the output is a valid JSON using the parameter \texttt{response\_format} in the \href{https://platform.openai.com/docs/api-reference/chat/create\#chat-create-response_format}{OpenAI API}. At the price of \$0.01 per 1k input tokens and \$0.03 per 1k generated tokens, the evaluation process costs approximately \$45 in total.

\subsection{Aligning the Errors}
\label{app:gpt4eval_matching}

For aligning the errors with the original text, we perform string matching on the text span decoded by GPT-4 in the \texttt{TEXT\_SPAN} field. In our preliminary experiments, this method proved to be more robust than either asking for start and end indices of the error span (which would rely on the model's ability to count characters) or performing sequence tagging on the copy of the input (which would rely on the model's ability to perfectly copy the input).

We tried to respect the monotonic ordering of text spans but fell back to full-text search if the span is not found following the previous one. We consider this approach successful since matching completely failed only in a minority of cases (137 out of 6927). Based on our manual examination, these mostly include cases where GPT-4 tried to suggest a \textit{missing} piece of text as an error or did not manage to copy the input text verbatim.

\section{Experiments with Open LLMs as Evaluators}
\label{app:openllmeval}
To select the most suitable LLM for automatic evaluation, we compared correlations with human judgment of the following models:

\begin{itemize}
  \item \textbf{GPT-4} \cite{openai2023gpt4} used via OpenAI API (\texttt{gpt-4-1106-preview}),
  \item \textbf{GPT-3.5} \cite{chatgpt} used via OpenAI API (\texttt{gpt-3.5-turbo-1106}),
  \item \textbf{Llama-3-70B}\footnote{\url{https://llama.meta.com/llama3/}} running locally via \href{https://ollama.com/}{Ollama} in 4-bit quantization (\texttt{meta-llama/Meta-Llama-3-70B}).
\end{itemize}

We used all the models with the same prompts, temperature 0, and force-decoded JSON outputs. In \autoref{tab:llmeval}, we show Pearson correlation coefficients computed as described in §\ref{sec:metricscorr}. We can see that the strongest model is GPT-4, followed by Llama-3-70B and GPT-3.5. As the gap between the models is substantial, we opted for using GPT-4 which is the strongest model.

\begin{table}[h]
  \centering
  \begin{tabular}{lccc} \toprule                                                             \\
    model      & $r_{\text{word}}$ & $r_{\text{example}}$ & $r_{\text{domain}}$ \\\midrule
    GPT-4      & \textbf{0.26}     & \textbf{0.52}        & \textbf{0.93}       \\
    GPT-3.5    & 0.07              & 0.33                 & 0.82                \\
    Llama3-70B & 0.09              & 0.44                 & 0.92                \\\bottomrule
  \end{tabular}
  \caption{Pearson correlation coefficients of the model annotations as compared with human annotations (cf. §\ref{sec:metricscorr}).}
  \label{tab:llmeval}
\end{table}


\section{Examples}
\label{app:outputs}
\begin{figure*}[h]
  \centering
  \small
  \includegraphics[width=\textwidth]{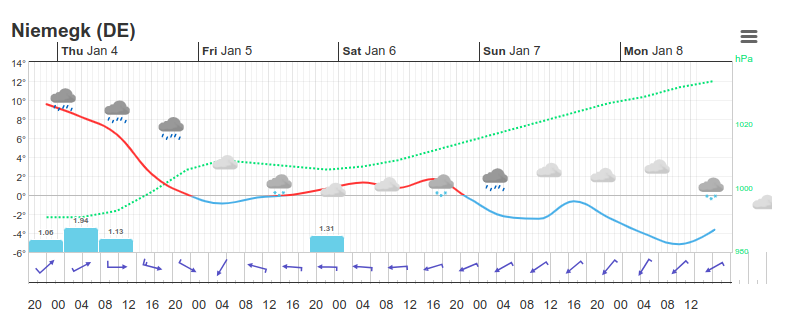}
  \caption{An example of an input from \texttt{openweather} (see the outputs in \autoref{tab:openweather}).}\label{fig:openweather}
\end{figure*}

\begin{figure*}[h]
  \centering
  \small
  \includegraphics[width=\textwidth]{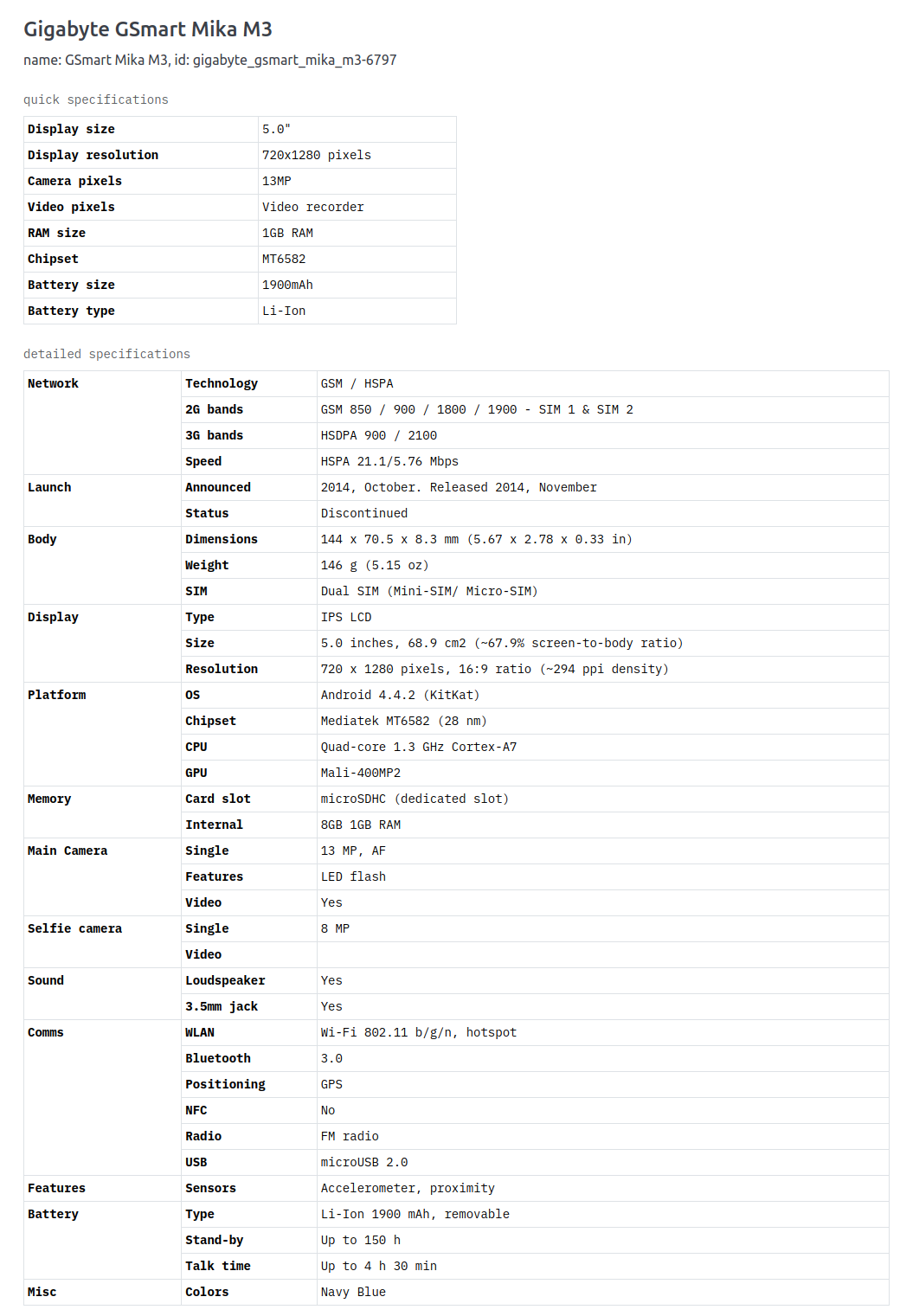}
  \caption{An example of an input from \texttt{gsmarena} (see the outputs in \autoref{tab:gsmarena}).}\label{fig:gsmarena}
\end{figure*}

\begin{figure*}[h]
  \centering
  \small
  \includegraphics[width=\textwidth]{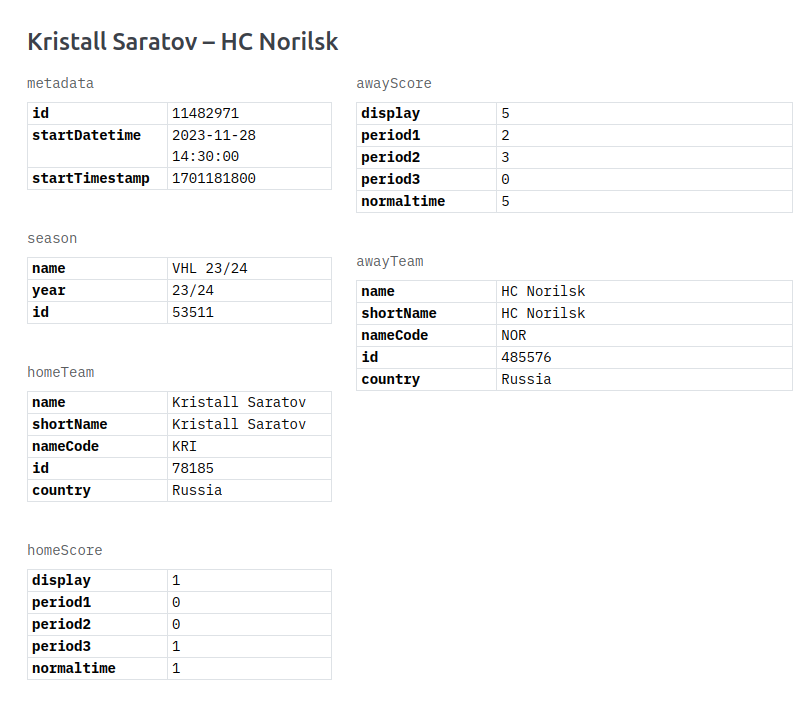}
  \caption{An example of an input from \texttt{ice\_hockey} (see the outputs in \autoref{tab:ice_hockey}).}\label{fig:ice_hockey}
\end{figure*}

\begin{figure*}[h]
  \centering
  \small
  \includegraphics[width=\textwidth]{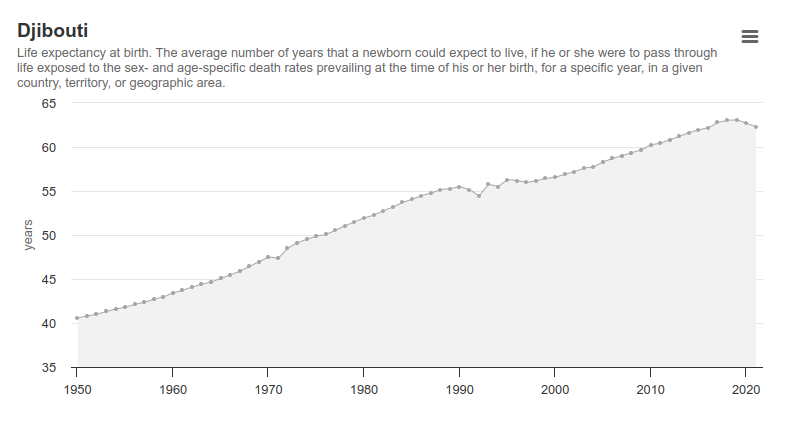}
  \caption{An example of an input from \texttt{owid} (see the outputs in \autoref{tab:owid}).}\label{fig:owid}
\end{figure*}

\begin{figure*}[h]
  \centering
  \small
  \includegraphics[width=0.8\textwidth]{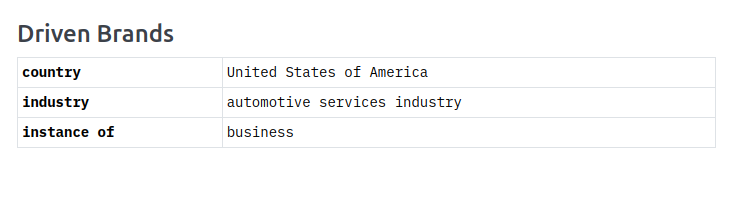}
  \caption{An example of an input from \texttt{wikidata} (see the outputs in \autoref{tab:wikidata}).}\label{fig:wikidata}
\end{figure*}

Here, we present an example of inputs and model outputs (along with annotations) for each domain:

\begin{itemize}
  \item \texttt{openweather}: \autoref{fig:openweather} (in) and \autoref{tab:openweather} (out),
  \item \texttt{gsmarena}: \autoref{fig:gsmarena} (in) and \autoref{tab:openweather} (out),
  \item \texttt{ice\_hockey}: \autoref{fig:ice_hockey} (in) and \autoref{tab:ice_hockey} (out),
  \item \texttt{owid}: \autoref{fig:owid} (in) and \autoref{tab:owid} (out),
  \item \texttt{wikidata}: \autoref{fig:wikidata} (in) and \autoref{tab:wikidata} (out).
\end{itemize}

Note that the graphs for \texttt{openweather} and \texttt{owid} are interactive when accessed through the web interface.

\section{Full Results}
\label{app:full_results}
Here, we include the tables with results for individual domains:

\begin{itemize}
  \item \autoref{tab:results_full} presents the average \textit{numbers of errors per output} separately for each domain (the aggregated results are in \autoref{tab:results_agg}),
  \item \autoref{tab:results_full_errperex} presents the ratio of \textit{outputs containing at least one error} separately for each domain (the aggregated results are in \autoref{tab:results_errperex}).
\end{itemize}

\begin{figure*}[h]
  \centering
  \small
  \textbf{System Message}
  \begin{verbatimbox}
    You are an expert data-to-text error annotation system. You undestand structured data and you can correcly operate with units and numerical values. You are designed to output token-level annotations in JSON.
  \end{verbatimbox}
  \textbf{Prompt}
  \begin{verbatimbox}
    Given the data:

    \`{}\`{}\`{}

    {data}

    \`{}\`{}\`{}

    Annotate all the errors in the following text:

    \`{}\`{}\`{}

    {text}

    \`{}\`{}\`{}

    Output the errors as a JSON list "errors" in which each object contains fields  "reason", "text", and "type". The value of "text" is the text of the error. The value of "reason" is the reason for the error. The value of "type" is one of {{0, 1, 2, 3}} based on the following list:

    - 0: Incorrect fact: The fact in the text contradicts the data.

    - 1: Not checkable: The fact in the text cannot be checked in the data.

    - 2: Misleading: The fact in the text is misleading in the given context.

    - 3: Other: The text is problematic for another reason, e.g. grammatically or stylistically incorrect, irrelevant, or repetitive.

    The list should be sorted by the position of the error in the text.

    *Example:*

    data:

    \`{}\`{}\`{}

    [ [ "Aditi Bhagwat", "occupation", "television actor" ], [ "Aditi Bhagwat", "date of birth", "18 January 1981" ] ]

    \`{}\`{}\`{}

    text:

    \`{}\`{}\`{}

    Aditi Bhagwat, born on January 18, 1991, used to be a popular Indian television actor. The data comes from a knowledge graph.

    \`{}\`{}\`{}

    output:

    \`{}\`{}\`{}{{ "errors": [{{"reason": "The data mentions that the actor was born on 1981", "text": "1991", "type": 0}}, {{"reason": "Misleadingly suggests that the actor is not alive", "text": "used to be", type: 2}}, {{"reason": "Popularity is not mentioned in the data", "text": "popular", type: 1}}, {{"reason", "Nationality is not mentioned in the data", "text": "Indian", type: 1}}, {{"reason": "The note is superfluous", "text": "The data comes from a knowledge graph.", type: 3}}] }}

    \`{}\`{}\`{}

    Note that some details may not be mentioned in the text: do not count omissions as errors. Also do not be too strict: some facts can be less specific than in the data (rounded values, shortened or abbreviated text, etc.), do not count these as errors. If there are no errors in the text, "errors" will be an empty list.
  \end{verbatimbox}

  \caption{The prompt we used for the GPT-4 evaluation metric.}
  \label{fig:gpt4prompt}
\end{figure*}

\begin{figure*}
  \footnotesize
  \begin{externaldoc}
    In this task, you will annotate \textbf{20} examples in total. For each example, you will see \textbf{data} on
    the left side and the corresponding generated \textbf{text} on the right
    side. Your task is to \textbf{annotate errors} in the text with respect
    to the data.

    There are five types of errors that you can mark in the generated text:

    \begin{enumerate}[nosep]
      \item
            \errinc{\textbf{Incorrect fact}}: The fact in the text contradicts the data.
      \item
            \errnc{\textbf{Not checkable}} : The fact in the text cannot be checked
            given the data.
      \item
            \errmis{\textbf{Misleading}}: The fact in the text is misleading in the given
            context.
      \item
            \errother{\textbf{Other}} : The text is problematic for another reason, e.g.
            grammatically or stylistically incorrect, irrelevant, or repetitive.
    \end{enumerate}

    \hypertarget{accordion-all}{}
    \hypertarget{how-to-mark-and-submit-the-annotations}{%
      \subsection*{How to mark and submit the
        annotations?}\label{how-to-mark-and-submit-the-annotations}}

    \hypertarget{collapseOne}{}
    Use your mouse to \textbf{highlight specific parts of the text}
    containing the errors. To switch between error categories, repeatedly
    click on the highlighted text (the last click removes the highlight).
    Note that highlighting from the right to left can work better for longer
    spans.

    Once you think you have marked all the errors present in the text, click
    the \textbf{Mark example as complete} button (you can still update
    the annotation later). You will be able to submit the annotations once
    they are all are marked as complete.

    \hypertarget{how-should-i-decide-on-the-errors}{%
      \subsection*{How should I decide on the
        errors?}\label{how-should-i-decide-on-the-errors}}

    \hypertarget{collapseTwo}{}
    \begin{itemize}[nosep]
      \item
            Each error span should include all the words related to the error (but
            nothing else).
      \item
            If you think the fact is probably true, but cannot be derived from the
            data, mark it as not checkable.
      \item
            If you are not really sure if the fact should be marked as an error,
            leave it unmarked.
    \end{itemize}

    \hypertarget{an-example-of-the-annotated-output}{%
      \subsection*{An example of the annotated
        output}\label{an-example-of-the-annotated-output}}

    \hypertarget{collapseSix}{}
    An example of the data input and the corresponding text annotated with
    errors:

    \vspace*{0.25cm}
    \textbf{data}

    \textbf{Nokia 3310}

    \begin{itemize}[nosep]
      \item
            \textbf{color:} black, blue, grey
      \item
            \textbf{display:} 320x240px
    \end{itemize}
    \vspace*{0.25cm}
    \textbf{text (product description)}

    Nokia 3310 is \errnc{produced in Finland} and features a \errinc{320x320 display}. It
    is \errmis{ available in black color.} \errother{The data seem to provide only partial
      information about the phone.}

    \vspace*{0.25cm}
    \textbf{explanation}
    \begin{itemize}[nosep]
      \item \textbf{produced in Finland}: The country where the phone is produced is not
            mentioned in the data.
      \item \textbf{320x320}: The data mentions that the display has resolution 320x240px.
      \item \textbf{available in black color}: Misleadingly suggests that the phone is not
            available in other colors.
      \item \textbf{The data seem to provide only partial information about the phone.}:
            The note is irrelevant for the phone description.
    \end{itemize}

    \emph{Note that for the sake of brevity, this particular example is
      based on a small data input and contains many errors, which may not
      always be the case.}

    \hypertarget{what-kinds-of-data-and-text-can-i-encounter}{%
      \subsection*{What kinds of data and text can I
        encounter?}\label{what-kinds-of-data-and-text-can-i-encounter}}

    \hypertarget{collapseThree}{}
    You can encounter the following kinds of texts:

    \begin{itemize}[nosep]
      \item
            a 5-day weather forecast generated from weather data,
      \item
            a description of a product generated from product specifications
      \item
            an ice hockey game report generated from information about the game,
      \item
            a caption of a health-related chart,
      \item
            a description of an entity (human, country, film, or company) based on
            its properties.
    \end{itemize}

    \hypertarget{what-is-the-source-of-the-data-and-the-texts}{%
      \subsection*{What is the source of the data and the
        texts?}\label{what-is-the-source-of-the-data-and-the-texts}}

    \hypertarget{collapseFour}{}
    The data is downloaded from public sources (\href{https://openweathermap.org/}{openweathermap.org},
    \href{https://www.gsmarena.com/}{gsmarena.com},
    \href{https://rapidapi.com/}{rapidapi.com},
    \href{https://ourworldindata.org/}{ourworldindata.com},
    \href{https://www.wikidata.org/}{wikidata.com}). The texts are generated
    by large language models.

    \hypertarget{what-is-the-purpose-of-the-research}{%
      \subsection*{What is the purpose of the
        research?}\label{what-is-the-purpose-of-the-research}}

    \hypertarget{collapseFive}{}
    The research is conducted by the NLP research group at Charles University in Prague.
    The research is a part of the ERC project
    \href{https://cordis.europa.eu/project/id/101039303}{Next-Generation Natural Language Generation}
    and was approved by the ethical committee
    of the project.
    \\
    \strut \\
    By continuing, you agree that the anonymous annotations will be made
    available to the research community to help evaluating and improving
    data-to-text generation systems based on large language models.

  \end{externaldoc}
  \caption{The instructions given to the human annotators.}
  \label{fig:humaninstructions}
\end{figure*}

\begin{figure*}[ht]
  \centering
  \includegraphics[width=\textwidth]{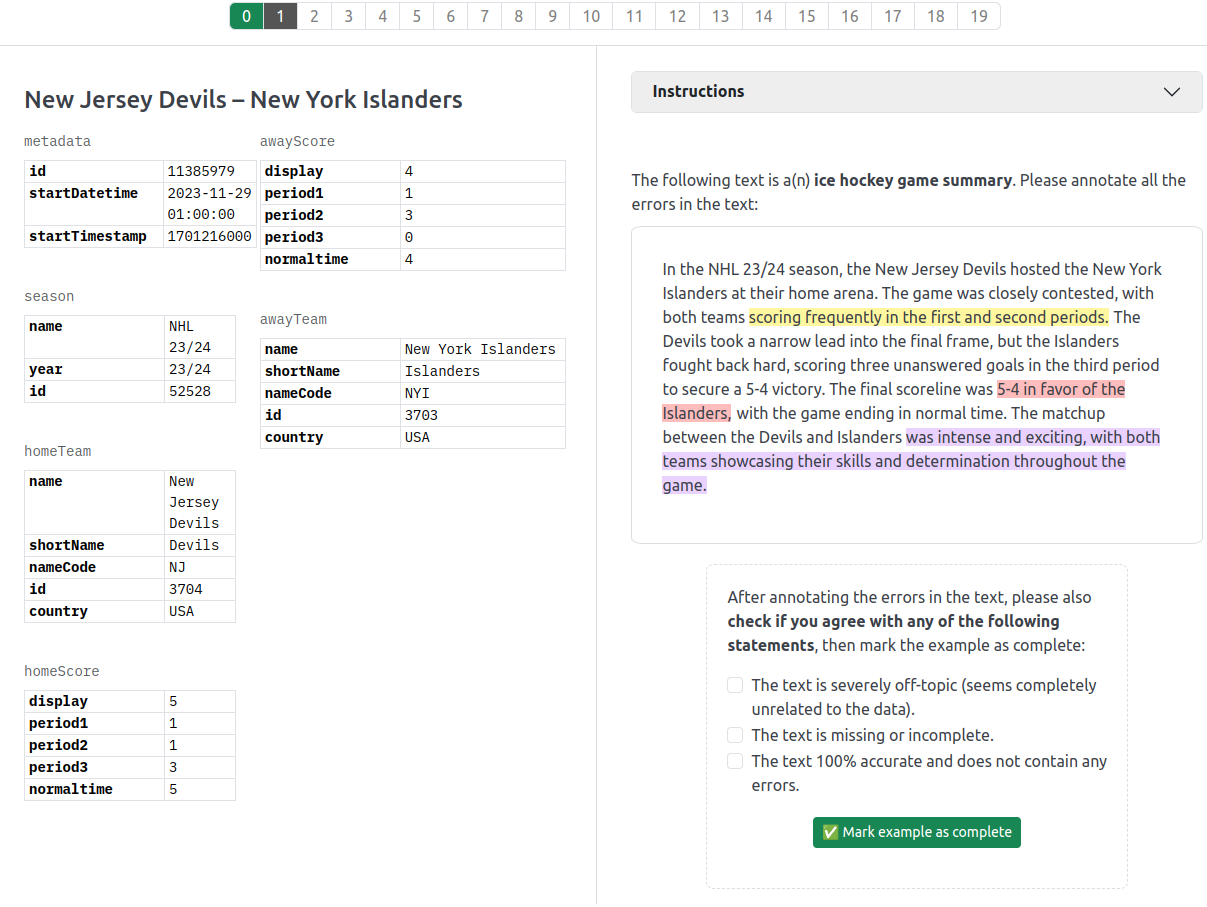}
  \caption{The annotation interface for human annotators.}\label{fig:interface}
\end{figure*}

\begin{table*}[ht]
  \small
  \begin{tabular}{lp{6.5cm}p{6.5cm}}
    \toprule
    \bfseries Model   & \textbf{Human annotations} (\humanmetric)                                                                                                                                                                                                                                                                                                                                                                                                                                                                                                                                                                                                                                                                                                                                                                                                                                                                                                                                                                                                                                                                                                                                                                                                                                                                                                                                                                                                                                                                                           & \textbf{GPT-4 annotations} (\gptmetric)                                                                                                                                                                                                                                                                                                                                                                                                                                                                                                                                                                                                                                                                                                                                                                                                                                                                                                                                                                                                                                                                                                                                                                                                                                                                                                                                                                                                                                                                  \\
    \midrule
    \bfseries Llama2  & The weather in Niemegk, Germany, will be cloudy and rainy for the next few days. On January 3rd, the temperature will be around 9.55\textdegree{}C, with a feels-like temperature of 6.59\textdegree{}C. The weather will be light rain, with a wind speed of 6.24 m/s and a gust of 12.1 m/s. \errnc{The rain will last for 3 hours}, and the pressure will be 991 hPa.                                                                                                                                                                                                                                                                                                                                                                                                                                                                                                                                                                                                                                                                                                                                                                                                                                                                                                                                                                                                                                                                                                                                                            & The weather in Niemegk, Germany, will be cloudy and rainy for the next few days. On January 3rd, the temperature will be around 9.55\textdegree{}C, with a feels-like temperature of 6.59\textdegree{}C. The weather will be light rain, with a wind speed of 6.24 m/s and a gust of 12.1 m/s. \errmis{The rain will last for 3 hours}, and the pressure will be 991 hPa.                                                                                                                                                                                                                                                                                                                                                                                                                                                                                                                                                                                                                                                                                                                                                                                                                                                                                                                                                                                                                                                                                                                     \\
    \bfseries Mistral & From January 3rd to January 7th, the city of Niemegk in Germany will experience a range of weather conditions. On January 3rd, the temperature will be 9.55\textdegree{}C with a feels like temperature of 6.59\textdegree{}C. There will be light rain with a \errnc{3-hour precipitation} of 1.06mm. The wind speed will be 6.24m/s with a gust of 12.1m/s. On January 4th, the temperature will be 8.2\textdegree{}C with a feels like temperature of 5.64\textdegree{}C. There will be light rain with a \errinc{3-hour precipitation} of 1.94mm. The wind speed will be 4.31m/s with a gust of 11.6m/s. On January 5th, the temperature will be \errinc{6.37\textdegree{}C with a feels like temperature of 2.25\textdegree{}C}. There will be light rain with a \errinc{3-hour precipitation of 1.13mm}. \errinc{The wind speed will be 7.07m/s with a gust of 15.51m/s}. On January 6th, the temperature will be \errinc{2.17\textdegree{}C} with a \errinc{feels like temperature of -3.55\textdegree{}C}. There will be overcast clouds with a \errinc{3-hour precipitation of 0.31mm}. The wind speed will be \errinc{3.54m/s with a gust of 7.11m/s}. On January 7th, the temperature will be -\errinc{0.89\textdegree{}C with a feels like temperature of -0.89\textdegree{}C}. There will be overcast clouds with a \errinc{3-hour precipitation of 0.31mm. The wind speed will be 0.63m/s with a gust of 0.7m/s}. & From January 3rd to January 7th, the city of Niemegk in Germany will experience a range of weather conditions. On January 3rd, the temperature will be 9.55\textdegree{}C with a feels like temperature of 6.59\textdegree{}C. There will be light rain with a 3-hour precipitation of 1.06mm. The wind speed will be 6.24m/s with a gust of 12.1m/s. On January 4th, the temperature will be 8.2\textdegree{}C with a feels like temperature of 5.64\textdegree{}C. There will be light rain with a 3-hour precipitation of 1.94mm. The wind speed will be 4.31m/s with a gust of 11.6m/s. On January 5th, the temperature will be 6.37\textdegree{}C with a feels like temperature of 2.25\textdegree{}C. There will be light rain with a 3-hour precipitation of \errinc{1.13mm}. The wind speed will be 7.07m/s with a gust of 15.51m/s. On January 6th, the temperature will be 2.17\textdegree{}C with a feels like temperature of -3.55\textdegree{}C. There will be overcast clouds with a 3-hour precipitation of \errinc{0.31mm}. The wind speed will be \errinc{3.54m/s} with a gust of \errinc{7.11m/s}. On January 7th, the temperature will be \errinc{-0.89\textdegree{}C} with a feels like temperature of \errinc{-0.89\textdegree{}C}. There will be overcast clouds with a 3-hour precipitation of \errinc{0.31mm}. The wind speed will be \errinc{0.63m/s} with a gust of \errinc{0.7m/s}. \\
    \bfseries Zephyr  & Over the next five days, the weather in Niemegk, Germany will be characterized by fluctuating temperatures and varying precipitation. On Thursday, January \errinc{3}, light rain and a \errinc{low} temperature of 9.55\textdegree{}C can be expected, with a chance of rain continuing into Friday. Snowfall is possible on Friday, with a low temperature of -0.89\textdegree{}C. Saturday will be overcast with a low temperature of \errinc{0.03\textdegree{}C}, and \errinc{light rain is forecasted} for Saturday evening. Sunday will be overcast as well, with a low temperature of -\errinc{2.23\textdegree{}C}. Monday will see a drop in temperatures, with a low of -\errinc{4.07\textdegree{}C} and scattered clouds. \errnc{Temperatures will continue to fall throughout the week, with lows of -5.19\textdegree{}C and -9.51\textdegree{}C on Tuesday and Wednesday, respectively}. Wind speeds will vary from 3.1 to \errinc{15.51} meters per second, with gusts reaching up to \errinc{15.51} meters per second on Monday.                                                                                                                                                                                                                                                                                                                                                                                                      & Over the next five days, the weather in Niemegk, Germany will be characterized by fluctuating temperatures and varying precipitation. On \errinc{Thursday, January 3}, light rain and a \errinc{low temperature of 9.55\textdegree{}C} can be expected, with a chance of rain continuing into Friday. Snowfall is possible on Friday, with a \errinc{low temperature of -0.89\textdegree{}C}. Saturday will be overcast with a low temperature of 0.03\textdegree{}C, and \errinc{light rain is forecasted for Saturday evening}. Sunday will be overcast as well, with a \errinc{low temperature of -2.23\textdegree{}C}. Monday will see a drop in temperatures, with a \errinc{low of -4.07\textdegree{}C} and scattered clouds. Temperatures will continue to fall throughout the week, with \errinc{lows of -5.19\textdegree{}C and -9.51\textdegree{}C on Tuesday and Wednesday}, respectively. Wind speeds will vary from 3.1 to 15.51 meters per second, with \errinc{gusts reaching up to 15.51 meters per second on Monday}.                                                                                                                                                                                                                                                                                                                                                                                   \\
    \bfseries GPT-3.5 & Over the next five days, the weather in Niemegk, Germany is expected to be quite varied. On January 4th, there will be light rain with temperatures ranging from \errinc{2.17\textdegree{}C to 9.55\textdegree{}C}. The following day, January 5th, there will be \errinc{light} snow with temperatures dropping to as low as -0.89\textdegree{}C. On January 6th, the weather will be overcast with temperatures ranging from -0.32\textdegree{}C to 1.69\textdegree{}C. The trend continues on January 7th with overcast clouds and temperatures dropping to as low as -\errinc{2.51\textdegree{}C}. Finally, on January 8th, there will be scattered clouds with temperatures ranging from -5.19\textdegree{}C to -3.67\textdegree{}C. Throughout this period, wind speeds will vary between 0.63 m/s and 8.12 m/s, and the pressure will range from \errinc{991} hPa to 1034 hPa.                                                                                                                                                                                                                                                                                                                                                                                                                                                                                                                                                                                                       & Over the next five days, the weather in Niemegk, Germany is expected to be quite varied. On January 4th, there will be light rain with temperatures ranging from 2.17\textdegree{}C to \errinc{9.55\textdegree{}C}. The following day, January 5th, there will be light snow with temperatures dropping to as low as \errinc{-0.89\textdegree{}C}. On January 6th, the weather will be overcast with temperatures ranging from -0.32\textdegree{}C to \errinc{1.69\textdegree{}C}. The trend continues on January 7th with overcast clouds and temperatures dropping to as low as \errinc{-2.51\textdegree{}C}. Finally, on January 8th, there will be scattered clouds with temperatures ranging from -5.19\textdegree{}C to \errinc{-3.67\textdegree{}C}. Throughout this period, wind speeds will vary between 0.63 m/s and \errinc{8.12 m/s}, and the pressure will range from \errinc{991 hPa} to \errinc{1034 hPa}.                                                                                                                                                                                                                                                                                                                                                                                                                                                                                                \\
    \bottomrule
\end{tabular}

  \caption{An example of the model outputs on \texttt{openweather} (see the input in \autoref{tab:openweather}).}\label{tab:openweather}
\end{table*}
\clearpage

\clearpage
\begin{table*}[h]
  \small
  \begin{tabular}{lp{6.5cm}p{6.5cm}}
    \toprule
    \bfseries Model   & \textbf{Human annotations} (\humanmetric)                                                                                                                                                                                                                                                                                                                                                                                                                                                                                                                                                                                                                                                                                                                                                                                                                                                                                                                                                                                 & \textbf{GPT-4 annotations} (\gptmetric)                                                                                                                                                                                                                                                                                                                                                                                                                                                                                                                                                                                                                                                                                                                                                                                                                                                                                                                                                                                                        \\
    \midrule
    \bfseries Llama2  & The Gigabyte GSmart Mika M3 is a smartphone that was released in 2014. It has a 5-inch IPS LCD display with a resolution of 720x1280 pixels and a 13MP main camera. The device also has an 8MP front camera and is powered by a Mediatek MT6582 chipset with 1GB of RAM and 8GB of internal storage. It has dual SIM support and is available in Navy Blue.                                                                                                                                                                                                                                                                                                                                                                                                                                                                                                                                                                                                                                                               & The Gigabyte GSmart Mika M3 is a smartphone that was released in 2014. It has a \errinc{5-inch} IPS LCD display with a resolution of 720x1280 pixels and a 13MP main camera. The device also has an \errnc{8MP front camera} and is powered by a Mediatek MT6582 chipset with 1GB of RAM and 8GB of internal storage. It has dual SIM support and \errnc{is available in Navy Blue}.                                                                                                                                                                                                                                                                                                                                                                                                                                                                                                                                                                                                                                                           \\
    \bfseries Mistral & The GSmart Mika M3 is a smartphone developed by Gigabyte. It was released in November 2014 and is powered by the Android 4.4.2 operating system. The device features a 5.0-inch IPS LCD display with a resolution of 720 x 1280 pixels, which translates to 294 pixels per inch. The smartphone is equipped with a 13-megapixel rear camera \errnc{with autofocus} and LED flash, as well as an 8-megapixel front-facing camera. It is powered by a Mediatek MT6582 chipset with a quad-core 1.3 GHz Cortex-A7 CPU and a Mali-400MP2 GPU. The device has 8GB of internal storage and 1GB of RAM, with the option to expand storage via a microSDHC card. The GSmart Mika M3 also features Wi-Fi, Bluetooth, GPS, and \errinc{NFC connectivity}, as well as a 3.5mm headphone jack and a microUSB 2.0 port for charging and data transfer. The device is available in Navy Blue and is powered by a 1900mAh Li-Ion battery, which provides up to 150 hours of stand-by time and up to 4 hours and 30 minutes of talk time. & The GSmart Mika M3 is a smartphone developed by Gigabyte. It was \errinc{released in November 2014} and is powered by the Android 4.4.2 operating system. The device features a 5.0-inch IPS LCD display with a resolution of 720 x 1280 pixels, which translates to 294 pixels per inch. The smartphone is equipped with a 13-megapixel rear camera with autofocus and LED flash, as well as an 8-megapixel front-facing camera. It is powered by a Mediatek MT6582 chipset with a quad-core 1.3 GHz Cortex-A7 CPU and a Mali-400MP2 GPU. The device has 8GB of internal storage and 1GB of RAM, with the option to expand storage via a microSDHC card. The GSmart Mika M3 also features Wi-Fi, Bluetooth, GPS, and \errinc{NFC connectivity}, as well as a 3.5mm headphone jack and a microUSB 2.0 port for charging and data transfer. The device is \errmis{available in Navy Blue} and is powered by a 1900mAh Li-Ion battery, which provides up to 150 hours of stand-by time and up to 4 hours and 30 minutes of talk time. \\
    \bfseries Zephyr  & The Gigabyte GSmart Mika M3 is a smartphone that boasts a 5.0-inch IPS LCD display with a resolution of 720 x 1280 pixels. It runs on Android 4.4.2 (KitKat) and is powered by a quad-core 1.3 GHz Cortex-A7 processor with a Mali-400MP2 GPU. The device comes with 8GB of internal storage and 1GB of RAM, which can be expanded via a dedicated microSDHC card slot. The primary camera is a 13-megapixel \errnc{autofocus lens} with an LED flash, while the secondary camera is an 8-megapixel lens. Connectivity options include Wi-Fi, Bluetooth, GPS, FM radio, and microUSB 2.0. The smartphone is equipped with a 1900mAh removable Li-Ion battery that provides up to \errinc{4 hours} of talk time and up to 150 hours of standby time. The device is available in Navy Blue and supports dual SIM cards (Mini-SIM/ Micro-SIM).                                                                                                                                                                               & The Gigabyte GSmart Mika M3 is a smartphone that boasts a 5.0-inch IPS LCD display with a resolution of 720 x 1280 pixels. It runs on Android 4.4.2 (KitKat) and is powered by a quad-core 1.3 GHz Cortex-A7 processor with a Mali-400MP2 GPU. The device comes with 8GB of internal storage and 1GB of RAM, which can be expanded via a dedicated microSDHC card slot. The primary camera is a 13-megapixel autofocus lens with an LED flash, while the secondary camera is an 8-megapixel lens. Connectivity options include Wi-Fi, Bluetooth, GPS, FM radio, and microUSB 2.0. The smartphone is equipped with a 1900mAh removable Li-Ion battery that provides \errinc{up to 4 hours of talk time} and up to 150 hours of standby time. The device is available in Navy Blue and supports dual SIM cards (Mini-SIM/ Micro-SIM).                                                                                                                                                                                                            \\
    \bfseries GPT-3.5 & The GSmart Mika M3, also known as the Gigabyte GSmart Mika M3, is a smartphone that was announced in October 2014 and released in November of the same year. It features a 5.0-inch IPS LCD display with a resolution of 720 x 1280 pixels and is powered by a Quad-core 1.3 GHz Cortex-A7 processor. The phone comes with 1GB of RAM and 8GB of internal storage, which can be expanded via a microSDHC card slot. The main camera is 13MP with LED flash, while the selfie camera is 8MP. The phone runs on Android 4.4.2 (KitKat) and is equipped with a 1900mAh removable battery. Other features include dual SIM capability, Wi-Fi, Bluetooth, GPS, FM radio, and microUSB 2.0. The GSmart Mika M3 is available in the color Navy Blue and has a sleek and compact design, measuring 144 x 70.5 x 8.3 mm and weighing 146g.                                                                                                                                                                                         & The GSmart Mika M3, also known as the \errinc{Gigabyte GSmart Mika M3}, is a smartphone that was announced in October 2014 and released in November of the same year. It features a 5.0-inch IPS LCD display with a resolution of 720 x 1280 pixels and is powered by a Quad-core 1.3 GHz Cortex-A7 processor. The phone comes with 1GB of RAM and 8GB of internal storage, which can be expanded via a microSDHC card slot. \errnc{The main camera is 13MP with LED flash}, \errnc{while the selfie camera is 8MP}. The phone runs on Android 4.4.2 (KitKat) \errnc{and is equipped with a 1900mAh removable battery}. Other features include dual SIM capability, Wi-Fi, Bluetooth, GPS, FM radio, and microUSB 2.0. \errnc{The GSmart Mika M3 is available in the color Navy Blue and has a sleek and compact design}, measuring 144 x 70.5 x 8.3 mm and weighing 146g.                                                                                                                                                                     \\
    \bottomrule
\end{tabular}

  \caption{An example of the model outputs on \texttt{gsmarena} (see the input in \autoref{tab:gsmarena}).}\label{tab:gsmarena}
\end{table*}
\clearpage


\begin{table*}[h]
  \small
  \begin{tabular}{lp{6.5cm}p{6.5cm}}
    \toprule
    \bfseries Model   & \textbf{Human annotations} (\humanmetric)                                                                                                                                                                                                                                                                                                                                                                                                                                                                                                       & \textbf{GPT-4 annotations} (\gptmetric)                                                                                                                                                                                                                                                                                                                                                                                                                                                                                                \\
    \midrule
    \bfseries Llama2  & In the VHL 23/24 season, Kristall Saratov hosted HC Norilsk at their home arena. Kristall Saratov \errinc{won the game} with a final score of 1-5. Kristall Saratov's home team \errinc{scored one goal in the first period}, \errinc{while HC Norilsk scored two goals in the second and third periods}. \errnc{The game lasted for 60 minutes}, with Kristall Saratov's home team \errinc{leading the game for the entire duration}.                                                                                                          & In the VHL 23/24 season, Kristall Saratov hosted HC Norilsk at their home arena. \errinc{Kristall Saratov won the game with a final score of 1-5.} \errinc{Kristall Saratov's home team scored one goal in the first period}, \errinc{while HC Norilsk scored two goals in the second and third periods.} \errnc{The game lasted for 60 minutes}, \errinc{with Kristall Saratov's home team leading the game for the entire duration.}                                                                                                 \\
    \bfseries Mistral & On November 28th, 2023, the VHL 23/24 season game between Kristall Saratov and HC Norilsk took place. The game was played at the home arena of Kristall Saratov, with the home team \errinc{winning 1-0}. The game was played in three periods, with the first period \errinc{ending in a 0-0 tie}, the second period ending in \errinc{a 2-3 score in favor of HC Norilsk}, and the third period ending in a 1-0 score in favor of Kristall Saratov. The game ended in normal time, with a \errinc{total score of 3-1 in favor of HC Norilsk}. & On November 28th, 2023, the VHL 23/24 season game between Kristall Saratov and HC Norilsk took place. The game was played at the home arena of Kristall Saratov, \errinc{with the home team winning 1-0}. The game was played in three periods, with the first period ending in a 0-0 tie, \errinc{the second period ending in a 2-3 score in favor of HC Norilsk}, and the third period ending in a 1-0 score in favor of Kristall Saratov. The game ended in normal time, \errinc{with a total score of 3-1 in favor of HC Norilsk}. \\
    \bfseries Zephyr  & In the VHL 23/24 season, Kristall Saratov hosted HC Norilsk. The first period saw two \errnc{quick} goals from Norilsk, with \errinc{one in the second and third periods as well}, resulting in a dominant \errinc{5-0} victory. Kristall Saratov managed to score a single goal in the third period, but it was not enough to overcome Norilsk's lead. The final score was 1-5 in favor of HC Norilsk.                                                                                                                                         & In the VHL 23/24 season, Kristall Saratov hosted HC Norilsk. The first period saw two quick goals from Norilsk, \errinc{with one in the second} and third periods as well, resulting in a dominant \errinc{5-0 victory}. Kristall Saratov managed to score a single goal in the third period, but it was not enough to overcome Norilsk's lead. The final score was 1-5 in favor of HC Norilsk.                                                                                                                                        \\
    \bfseries GPT-3.5 & In a match from the VHL 23/24 season, Kristall Saratov faced off against HC Norilsk. The game took place on November 28, 2023, at \errinc{15:30}. Despite a goal from Kristall Saratov in the third period, HC Norilsk emerged victorious with a final score of 5-1. The away team dominated the game, scoring 2 goals in the first period and 3 goals in the second period. Kristall Saratov struggled to keep up, resulting in a clear win for HC Norilsk.                                                                                    & In a match from the VHL 23/24 season, Kristall Saratov faced off against HC Norilsk. The game took place on November 28, 2023, at 15:30. Despite a goal from Kristall Saratov in the third period, HC Norilsk emerged victorious with a final score of 5-1. The away team dominated the game, scoring 2 goals in the first period and 3 goals in the second period. Kristall Saratov struggled to keep up, resulting in a clear win for HC Norilsk.                                                                                    \\
    \bottomrule
\end{tabular}

  \caption{An example of the model outputs on \texttt{ice\_hockey} (see the input in \autoref{tab:ice_hockey}).}\label{tab:ice_hockey}
\end{table*}
\clearpage


\begin{table*}[h]
  \small
  \begin{tabular}{lp{6.5cm}p{6.5cm}}
    \toprule
    \bfseries Model   & \textbf{Human annotations} (\humanmetric)                                                                                                                                                                                                                                                                                                                                                                                                                                                                                                                                         & \textbf{GPT-4 annotations} (\gptmetric)                                                                                                                                                                                                                                                                                                                                                                                                                                                                                                                                                             \\
    \midrule
    \bfseries Llama2  & Life expectancy at birth in Djibouti has increased significantly over the past few decades, with a notable increase from 40.5665 years in 1950 to 62.3049 years in 2021.                                                                                                                                                                                                                                                                                                                                                                                                          & Life expectancy at birth in Djibouti has increased significantly over the past few decades, with a notable increase from 40.5665 years in 1950 to 62.3049 years in 2021.                                                                                                                                                                                                                                                                                                                                                                                                                            \\
    \bfseries Mistral & This chart displays the life expectancy at birth in Djibouti from 1950 to 2021. The data shows a steady increase in life expectancy over the years, with \errinc{a significant jump in the 1970s and 1980s}. In 2021, the average life expectancy at birth in Djibouti was 62.3 years.                                                                                                                                                                                                                                                                                            & This chart displays the life expectancy at birth in Djibouti from 1950 to 2021. The data shows a \errinc{steady increase} in life expectancy over the years, with a \errinc{significant jump in the 1970s and 1980s}. In 2021, the average life expectancy at birth in Djibouti was \errinc{62.3 years}.                                                                                                                                                                                                                                                                                            \\
    \bfseries Zephyr  & This line graph illustrates the evolution of life expectancy at birth in Djibouti from 1950 to 2021. Initially, the average number of years that a newborn could expect to live was around 40.5 years in 1950, and it gradually increased over the following decades, reaching 62.3 years in 2021. The trend shows a steady improvement, with some fluctuations in the \errinc{1980s} and 1990s. The data presented here is based on the sex- and age-specific death rates prevailing at the time of birth for a specific year in a given country, territory, or geographic area. & This line graph illustrates the evolution of life expectancy at birth in Djibouti from 1950 to 2021. Initially, the average number of years that a newborn could expect to live was \errinc{around 40.5 years in 1950}, and it gradually increased over the following decades, reaching \errinc{62.3 years in 2021}. \errmis{The trend shows a steady improvement}, with some fluctuations in the 1980s and 1990s. The data presented here is based on the sex- and age-specific death rates prevailing at the time of birth for a specific year in a given country, territory, or geographic area. \\
    \bfseries GPT-3.5 & The chart shows the life expectancy at birth in Djibouti from 1950 to 2021. The life expectancy has steadily increased over the years, from 40.57 years in 1950 to 62.30 years in 2021, with some fluctuations in between. The data reflects the average number of years a newborn could expect to live if exposed to the prevailing death rates at the time of birth.                                                                                                                                                                                                            & The chart shows the life expectancy at birth in Djibouti from 1950 to 2021. \errinc{The life expectancy has steadily increased over the years}, from \errinc{40.57 years in 1950} to \errinc{62.30 years in 2021}, with some fluctuations in between. The data reflects the average number of years a newborn could expect to live if exposed to the prevailing death rates at the time of birth.                                                                                                                                                                                                   \\
    \bottomrule
\end{tabular}

  \caption{An example of the model outputs on \texttt{owid} (see the input in \autoref{tab:owid}).}\label{tab:owid}
\end{table*}

\clearpage

\begin{table*}[h]
  \small
  \begin{tabular}{lp{6.5cm}p{6.5cm}}
    \toprule
    \bfseries Model   & \textbf{Human annotations} (\humanmetric)                                                                                                                                                                                                                                                                                                 & \textbf{GPT-4 annotations} (\gptmetric)                                                                                                                                                                                                                                                                                           \\
    \midrule
    \bfseries Llama2  & Driven Brands is an American \errnc{franchise holding} company that \errnc{owns, operates, and franchises} automotive services businesses in the United States. \errnc{The company was founded in 1996 and is headquartered in Charlotte, North Carolina. It operates several brands, including Maaco, Meineke, Econo Lube, and CARSTAR}. & Driven Brands is an American franchise holding company that owns, operates, and franchises automotive services businesses in the United States. The company was \errnc{founded in 1996} and is \errnc{headquartered in Charlotte, North Carolina}. It operates several brands, including Maaco, Meineke, Econo Lube, and CARSTAR. \\
    \bfseries Mistral & Driven Brands is a business operating in the automotive services industry in the United States of America.                                                                                                                                                                                                                                & Driven Brands is a business operating in the automotive services industry in the United States of America.                                                                                                                                                                                                                        \\
    \bfseries Zephyr  & Driven Brands is a business operating in the automotive services industry based in the United States of America. As an instance of the broader category of businesses, Driven Brands \errnc{provides goods or services to customers in exchange for financial remuneration}.                                                              & Driven Brands is a business operating in the automotive services industry based in the United States of America. \errother{As an instance of the broader category of businesses,} Driven Brands provides \errinc{goods or services} to customers in exchange for financial remuneration.                                          \\
    \bfseries GPT-3.5 & Driven Brands is a business based in the United States of America, operating within the automotive services industry.                                                                                                                                                                                                                     & Driven Brands is a business based in the United States of America, operating within the automotive services industry.                                                                                                                                                                                                             \\
    \bottomrule
\end{tabular}

  \caption{An example of the model outputs on \texttt{wikidata} (see the input in \autoref{tab:wikidata}).}\label{tab:wikidata}
\end{table*}

\clearpage

\begin{table*}[t]
  \small
  \centering
  \begin{tabular}{@{}clccccccccccr@{}}
    \toprule
    \multicolumn{1}{l}{}              &                      & \multicolumn{2}{c}{\textbf{Incorrect}} & \multicolumn{2}{c}{\textbf{Not Checkable}} & \multicolumn{2}{c}{\textbf{Misleading}} & \multicolumn{2}{c}{\textbf{Other}} & \multicolumn{2}{c}{\textbf{All categories}} &                                                                                                 \\
    \multicolumn{1}{l}{}              &                      & \humanmetric{}                         & \gptmetric{}                               & \humanmetric{}                          & \gptmetric{}                       & \humanmetric{}                              & \gptmetric{} & \humanmetric{} & \gptmetric{} & \humanmetric{} & \gptmetric{} & \# \textbf{Tok.} \\\midrule

    \multirow{5}{*}{\textbf{Llama 2}} & \texttt{openweather} & 4.27                                   & 6.99                                       & 2.23                                    & 1.93                               & 0.58                                        & 0.20         & 0.16           & 0.26         & 7.24           & 9.38         & 166.7            \\
                                      & \texttt{gsmarena}    & 0.67                                   & 1.96                                       & 0.87                                    & 0.31                               & 0.20                                        & 0.24         & 0.09           & 0.10         & 1.83           & 2.61         & 90.3             \\
                                      & \texttt{ice\_hockey} & 1.47                                   & 1.98                                       & 0.99                                    & 0.90                               & 0.09                                        & 0.00         & 0.12           & 0.05         & 2.67           & 2.93         & 63.3             \\
                                      & \texttt{owid}        & 1.30                                   & 2.26                                       & 0.58                                    & 0.15                               & 0.32                                        & 0.13         & 0.07           & 0.00         & 2.27           & 2.54         & 60.3             \\
                                      & \texttt{wikidata}    & 0.15                                   & 0.74                                       & 1.59                                    & 1.24                               & 0.08                                        & 0.03         & 0.07           & 0.04         & 1.89           & 2.05         & 38.5             \\ \cdashlinelr{1-13}
    \multirow{5}{*}{\textbf{Mistral}} & \texttt{openweather} & 5.65                                   & 8.76                                       & 2.47                                    & 0.41                               & 1.36                                        & 0.50         & 0.53           & 0.04         & 10.01          & 9.71         & 193.5            \\
                                      & \texttt{gsmarena}    & 0.51                                   & 2.02                                       & 1.10                                    & 0.94                               & 0.30                                        & 0.25         & 0.14           & 0.02         & 2.05           & 3.23         & 146.3            \\
                                      & \texttt{ice\_hockey} & 1.47                                   & 1.79                                       & 0.92                                    & 0.76                               & 0.14                                        & 0.08         & 0.13           & 0.10         & 2.66           & 2.73         & 92.4             \\
                                      & \texttt{owid}        & 2.42                                   & 3.11                                       & 0.40                                    & 0.18                               & 0.29                                        & 0.27         & 0.07           & 0.02         & 3.18           & 3.58         & 91.1             \\
                                      & \texttt{wikidata}    & 0.10                                   & 0.45                                       & 0.72                                    & 0.40                               & 0.13                                        & 0.19         & 0.38           & 0.31         & 1.33           & 1.35         & 51.0             \\ \cdashlinelr{1-13}
    \multirow{5}{*}{\textbf{Zephyr}}  & \texttt{openweather} & 4.22                                   & 7.39                                       & 1.01                                    & 0.37                               & 0.34                                        & 0.41         & 0.16           & 0.00         & 5.73           & 8.17         & 130.9            \\
                                      & \texttt{gsmarena}    & 0.33                                   & 2.24                                       & 1.04                                    & 1.00                               & 0.23                                        & 0.48         & 0.11           & 0.01         & 1.71           & 3.73         & 142.8            \\
                                      & \texttt{ice\_hockey} & 0.89                                   & 1.57                                       & 0.61                                    & 0.20                               & 0.11                                        & 0.04         & 0.07           & 0.01         & 1.68           & 1.82         & 83.1             \\
                                      & \texttt{owid}        & 1.68                                   & 2.64                                       & 0.49                                    & 0.16                               & 0.22                                        & 0.30         & 0.10           & 0.01         & 2.49           & 3.11         & 85.2             \\
                                      & \texttt{wikidata}    & 0.09                                   & 0.36                                       & 0.72                                    & 0.29                               & 0.12                                        & 0.21         & 0.36           & 0.21         & 1.29           & 1.07         & 48.1             \\ \cdashlinelr{1-13}
    \multirow{5}{*}{\textbf{GPT-3.5}} & \texttt{openweather} & 1.57                                   & 4.34                                       & 0.57                                    & 0.15                               & 0.38                                        & 0.53         & 0.05           & 0.00         & 2.57           & 5.02         & 112.8            \\
                                      & \texttt{gsmarena}    & 0.20                                   & 1.64                                       & 0.80                                    & 1.42                               & 0.21                                        & 0.34         & 0.17           & 0.01         & 1.38           & 3.41         & 129.5            \\
                                      & \texttt{ice\_hockey} & 0.81                                   & 0.76                                       & 0.46                                    & 0.17                               & 0.07                                        & 0.04         & 0.09           & 0.01         & 1.43           & 0.98         & 84.4             \\
                                      & \texttt{owid}        & 0.64                                   & 1.87                                       & 0.25                                    & 0.02                               & 0.17                                        & 0.27         & 0.01           & 0.00         & 1.07           & 2.16         & 62.2             \\
                                      & \texttt{wikidata}    & 0.05                                   & 0.18                                       & 0.36                                    & 0.16                               & 0.06                                        & 0.10         & 0.05           & 0.10         & 0.52           & 0.54         & 35.7             \\\bottomrule
  \end{tabular}
  \caption{The average \textit{numbers of errors per output} for each domain (lower is better). We also include the average number of tokens per output in the rightmost column. See \autoref{tab:results_agg} for aggregated results.}
  \label{tab:results_full}
\end{table*}

\begin{table*}[t]
  \small
  \centering
  \begin{tabular}{@{}clrrrrrrrrrr@{}}
    \toprule
                                      &                      & \multicolumn{2}{c}{\textbf{Incorrect}} & \multicolumn{2}{c}{\textbf{Not Checkable}} & \multicolumn{2}{c}{\textbf{Misleading}} & \multicolumn{2}{c}{\textbf{Other}} & \multicolumn{2}{c}{\textbf{All categories}}                                                                                \\
                                      &                      & \humanmetric{}                         & \gptmetric{}                               & \humanmetric{}                          & \gptmetric{}                       & \humanmetric{}                              & \gptmetric{} & \humanmetric{} & \gptmetric{} & \humanmetric{} & \gptmetric{} \\\midrule

    \multirow{5}{*}{\textbf{Llama 2}} & \texttt{openweather} & 76\%                                   & 98\%                                       & 71\%                                    & 68\%                               & 30\%                                        & 14\%         & 12\%           & 22\%         & 90\%           & 100\%        \\
                                      & \texttt{gsmarena}    & 38\%                                   & 86\%                                       & 49\%                                    & 24\%                               & 18\%                                        & 16\%         & 5\%            & 7\%          & 74\%           & 94\%         \\
                                      & \texttt{ice\_hockey} & 70\%                                   & 77\%                                       & 51\%                                    & 55\%                               & 7\%                                         & 0\%          & 10\%           & 5\%          & 89\%           & 94\%         \\
                                      & \texttt{owid}        & 70\%                                   & 92\%                                       & 37\%                                    & 14\%                               & 24\%                                        & 12\%         & 6\%            & 0\%          & 88\%           & 93\%         \\
                                      & \texttt{wikidata}    & 12\%                                   & 47\%                                       & 79\%                                    & 63\%                               & 8\%                                         & 2\%          & 5\%            & 4\%          & 87\%           & 89\%         \\
    \multirow{5}{*}{\textbf{Mistral}} & \texttt{openweather} & 78\%                                   & 99\%                                       & 61\%                                    & 16\%                               & 38\%                                        & 18\%         & 17\%           & 4\%          & 92\%           & 100\%        \\
                                      & \texttt{gsmarena}    & 33\%                                   & 89\%                                       & 54\%                                    & 54\%                               & 19\%                                        & 17\%         & 12\%           & 2\%          & 73\%           & 99\%         \\
                                      & \texttt{ice\_hockey} & 74\%                                   & 83\%                                       & 62\%                                    & 52\%                               & 12\%                                        & 8\%          & 9\%            & 7\%          & 88\%           & 97\%         \\
                                      & \texttt{owid}        & 75\%                                   & 96\%                                       & 27\%                                    & 8\%                                & 24\%                                        & 25\%         & 6\%            & 2\%          & 86\%           & 97\%         \\
                                      & \texttt{wikidata}    & 8\%                                    & 34\%                                       & 44\%                                    & 29\%                               & 10\%                                        & 17\%         & 24\%           & 27\%         & 67\%           & 72\%         \\
    \multirow{5}{*}{\textbf{Zephyr}}  & \texttt{openweather} & 82\%                                   & 99\%                                       & 45\%                                    & 25\%                               & 23\%                                        & 25\%         & 14\%           & 0\%          & 90\%           & 100\%        \\
                                      & \texttt{gsmarena}    & 22\%                                   & 91\%                                       & 47\%                                    & 52\%                               & 19\%                                        & 30\%         & 9\%            & 1\%          & 61\%           & 99\%         \\
                                      & \texttt{ice\_hockey} & 54\%                                   & 78\%                                       & 39\%                                    & 17\%                               & 11\%                                        & 4\%          & 6\%            & 1\%          & 76\%           & 87\%         \\
                                      & \texttt{owid}        & 70\%                                   & 93\%                                       & 31\%                                    & 9\%                                & 20\%                                        & 26\%         & 7\%            & 1\%          & 85\%           & 95\%         \\
                                      & \texttt{wikidata}    & 6\%                                    & 29\%                                       & 49\%                                    & 22\%                               & 8\%                                         & 18\%         & 22\%           & 18\%         & 66\%           & 66\%         \\
    \multirow{5}{*}{\textbf{GPT-3.5}} & \texttt{openweather} & 64\%                                   & 98\%                                       & 32\%                                    & 12\%                               & 23\%                                        & 25\%         & 5\%            & 0\%          & 75\%           & 99\%         \\
                                      & \texttt{gsmarena}    & 15\%                                   & 74\%                                       & 42\%                                    & 58\%                               & 13\%                                        & 19\%         & 12\%           & 1\%          & 57\%           & 97\%         \\
                                      & \texttt{ice\_hockey} & 64\%                                   & 50\%                                       & 30\%                                    & 12\%                               & 7\%                                         & 4\%          & 8\%            & 1\%          & 76\%           & 57\%         \\
                                      & \texttt{owid}        & 43\%                                   & 89\%                                       & 13\%                                    & 2\%                                & 14\%                                        & 24\%         & 1\%            & 0\%          & 57\%           & 90\%         \\
                                      & \texttt{wikidata}    & 4\%                                    & 14\%                                       & 27\%                                    & 14\%                               & 6\%                                         & 9\%          & 5\%            & 9\%          & 38\%           & 36\%         \\\bottomrule
  \end{tabular}
  \caption{The percentage of \textit{outputs containing at least one error} for each domain (lower is better). See \autoref{tab:results_errperex} for aggregated results.}
  \label{tab:results_full_errperex}
\end{table*}

\end{document}